\documentclass{article} %
\usepackage{iclr2024_conference,times}

\usepackage{amsmath,amsfonts,bm}

\def\eqref#1{equation~\ref{#1}}

\def\1{\bm{1}}

\def\rvg{{\mathbf{g}}}

\def\rvm{{\mathbf{m}}}

\def\rvv{{\mathbf{v}}}
\def\rvw{{\mathbf{w}}}
\def\rvx{{\mathbf{x}}}
\def\rvy{{\mathbf{y}}}
\def\rvz{{\mathbf{z}}}

\def\rmB{{\mathbf{B}}}
\def\rmC{{\mathbf{C}}}

\def\rmH{{\mathbf{H}}}
\def\rmI{{\mathbf{I}}}

\def\rmM{{\mathbf{M}}}

\def\rmP{{\mathbf{P}}}

\def\rmR{{\mathbf{R}}}
\def\rmS{{\mathbf{S}}}
\def\rmT{{\mathbf{T}}}

\def\mF{{\bm{F}}}

\def\mH{{\bm{H}}}

\def\mK{{\bm{K}}}

\def\mQ{{\bm{Q}}}
\def\mR{{\bm{R}}}

\DeclareMathAlphabet{\mathsfit}{\encodingdefault}{\sfdefault}{m}{sl}
\SetMathAlphabet{\mathsfit}{bold}{\encodingdefault}{\sfdefault}{bx}{n}

\DeclareMathOperator*{\argmax}{arg\,max}
\DeclareMathOperator*{\argmin}{arg\,min}

\usepackage{hyperref}
\usepackage{booktabs} %
\usepackage{url}
\usepackage{graphicx}
\usepackage{amsmath}
\usepackage{amssymb}
\usepackage{cleveref}
\usepackage{algorithm}
\usepackage{algorithmic}

\crefformat{equation}{#2(#1)#3}
\let\eqref\cref
\crefformat{algorithm}{#2Algorithm~#1#3}

\title{Implicit Maximum a Posteriori Filtering via Adaptive Optimization}

\author{Gianluca M. Bencomo$^1$, Jake C. Snell$^1$ \& Thomas L. Griffiths$^{1,2}$ \\ %
$^1$Department of Computer Science, Princeton University \\
$^2$Department of Psychology, Princeton University\\
\texttt{\{gb5435,js2523,tomg\}@princeton.edu} \\
}

\iclrfinalcopy %
\begin{document}

\maketitle

\begin{abstract}
Bayesian filtering approximates the true underlying behavior of a time-varying system by inverting an explicit generative model to convert noisy measurements into state estimates. This process typically requires either storage, inversion, and multiplication of large matrices or Monte Carlo estimation, neither of which are practical in high-dimensional state spaces such as the weight spaces of artificial neural networks. Here, we frame the standard Bayesian filtering problem as optimization over a time-varying objective. Instead of maintaining matrices for the filtering equations or simulating particles, we specify an optimizer that defines the Bayesian filter \emph{implicitly}. In the linear-Gaussian setting, we show that every Kalman filter has an equivalent formulation using $K$ steps of gradient descent. In the nonlinear setting, our experiments demonstrate that our framework results in filters that are effective, robust, and scalable to high-dimensional systems, comparing well against the standard toolbox of Bayesian filtering solutions. We suggest that it is easier to fine-tune an optimizer than it is to specify the correct filtering equations, making our framework an attractive option for high-dimensional filtering problems.
\end{abstract}

\section{Introduction}\label{sec:intro}

Time-varying systems are ubiquitous in science, engineering, and machine learning. Consider a multielectrode array receiving raw voltage signals from thousands of neurons during a visual perception task. The goal is to infer some underlying neural state that is not directly observable, such that we can draw connections between neural activity and visual perception, but raw voltage signals are a sparse representation of neural activity that is shrouded in noise. To confound the problem further, the underlying neural state changes throughout time in both expected and unexpected ways. This problem, and most time-varying prediction problems, can be formalized as a probablistic state space model where latent variables evolve over time and emit observations \citep{simon2006optimal}. One solution to such a problem is to apply a Bayesian filter, a type of probabilistic model that can infer the values of latent variables from observations.

Typically, Bayesian filters require matrix storage, inversion, and multiplication or the storage of \emph{particles}, which are samples from the filtering distribution at every time step. In large state and observation spaces, the computational costs associated with both of these approaches render them impractical \citep{raitoharju2019computational}. In addition, since most time-varying systems do not have ground-truth states available, accurately estimating the process noise -- the variability of the underlying latent variables -- is nearly impossible. This problem is exacerbated in continuous systems where we need to perform numerical integration, which introduces additional uncertainty. Existing Bayesian filters are sensitive to process noise and dynamics misspecification, making them hard to use in these settings \citep{mehra1972approaches}.  Most filters are highly structured, requiring explicit specification of distributional assumptions. To be usable in practical settings, we need a system that remains effective when scaled to large state spaces (being applicable to weather systems, neural recording, or even modeling the evolving weights of an artificial neural network) and is robust to misspecification of the process noise and dynamics (which is almost guaranteed). %

The practical problems that arise when training deep neural networks (DNNs) are not dissimilar from those that make Bayesian filters difficult to work with: high-dimensional state spaces, nonlinearities, and nonconvexity. These pressures have made way for adaptive optimizers \citep{duchi2011adaptive, tieleman2012lecture, zeiler2012adadelta, kingma2014adam} that offer a largely prescriptive solution for training DNNs, succeeding many previous approaches to optimizing neural networks that include applying Bayesian filtering equations \citep{haykin2004kalman}. To efficiently and effectively overcome the difficulties of training DNNs  \citep{glorot2010understanding}, modern optimizers invariably make use of (i) the empirical Fisher information to crudely incorporate curvature  at very little computational cost \citep{martens2020new} and (ii) momentum to increase stability and the rate of convergence \citep{sutskever2013importance}. We show that these same methods can be applied to Bayesian filtering in an extremely practical and theoretically-motivated framework.

In the following sections, we cast Bayesian filtering as optimization over a time-varying objective and show that much of the information that is explicitly defined by structured filters can be \emph{implicitly} internalized by optimizer hyperparameters. In doing so, we can solve many of the scalability challenges associated with Bayesian filters and make Bayesian filtering easier to implement, especially for the well-versed deep learning practitioner. We show that our proposed method, Implicit Maximum a Posteriori (IMAP) Filtering, is robust to misspecification and matches or outperforms classical filters on baseline tasks. We also show that it naturally scales up to high-dimensional problems such as adapting the weights of a convolutional neural network, and that it performs particularly well in this setting. We argue that it is easier to specify an optimizer than it is to correctly identify classical filtering equations, and our results illustrate the benefits of this implicit filtering approach.

\section{Background}\label{sec:background}
We begin with an introduction to the filtering problem. Assume a discrete-time state space model with a Markov process over the states $\rvx_{0:T} \triangleq \rvx_0, \rvx_1, \ldots, \rvx_T$ and a sequence of observations $\rvy_{1:T} \triangleq \rvy_1, \rvy_2, \ldots, \rvy_T$ which are conditionally independent given the corresponding state. The joint probability of the states and observations is
\begin{align}
    p(\rvx_{0:T}, \rvy_{1:T}) = p(\rvx_0) \prod_{t=1}^T p(\rvx_t | \rvx_{t-1}) p(\rvy_t | \rvx_t),
\end{align}
where $p(\rvx_0)$ is the \emph{initial distribution}, $p(\rvx_t | \rvx_{t-1})$ is the \emph{transition distribution}, and $p(\rvy_t | \rvx_t)$ is the \emph{likelihood}. The posterior over the joint state  $\rvx_{1:t}$ of the system given all of the observations up until the current time $\rvy_{1:t}$ is given by:
\begin{align}
    p(\rvx_{1:t} | \rvy_{1:t}) = \int p(\rvx_0) \prod_{s=1}^t \frac{p(\rvx_s | \rvx_{s-1}) p(\rvy_s | \rvx_s)}{p(\rvy_s | \rvy_{1:s-1})} \,d \rvx_0,
\end{align}
where $p(\rvy_1 | \rvy_{1:0}) \triangleq p(\rvy_1)$. In sequential estimation, we are often only interested in the marginal of the current state $p(\rvx_t | \rvy_{1:t})$, known as the \emph{filtering distribution}. We normally want to compute or approximate this distribution such that the requisite computational resources do not depend on the length of the sequence. This can be accomplished by first initializing $p(\rvx_0 | \rvy_{1:0}) \triangleq p(\rvx_0)$, and then forming a \emph{predictive distribution} via the Chapman-Kolmogorov equation,
\begin{align}
    p(\rvx_t|\rvy_{1:t-1}) = \int p(\rvx_t|\rvx_{t-1})p(\rvx_{t-1}|\rvy_{1:t-1}) d \rvx_{t-1}, \label{eq:chapman_kolmogorov}
\end{align}
and updating that predictive distribution, in light of measurements $\rvy_t$, by a simple application of Bayes' rule for every time step $t = 1, \ldots, T$,
\begin{align}
    p(\rvx_t|\rvy_{1:t}) = \frac{p(\rvy_t|\rvx_t)p(\rvx_t|\rvy_{1:t-1})}{p(\rvy_{t}|\rvy_{1:t-1})}.\label{eq:update_step}
\end{align}

\subsection{Kalman Filter for Linear-Gaussian Systems}
When the transition distribution and likelihood are linear and Gaussian,  we can compute optimal estimates in closed-form via the Kalman filtering equations \citep{kalman1960new}.
Assume the following state space model:
\begin{align}
    p(\rvx_0) &= \mathcal{N}(\rvx_0 \mid \bm{\mu}_0, \bm{\Sigma}_0), \\
    p(\rvx_t \mid \rvx_{t-1}) &= \mathcal{N}(\rvx_t \mid \mF_{t-1} \rvx_{t-1}, \mQ_{t-1}), \\
    p(\rvy_t \mid \rvx_t) &= \mathcal{N}(\rvy_t \mid \rmH_t \rvx_t, \rmR_t),
\end{align}
where $\mF_{t-1}$ and $\rmH_t$ are the transition and observation matrices, respectively, with process noise $\mQ_{t-1}$ and measurement noise $\rmR_t$.
The Kalman filter propagates the filtering distribution from the previous timestep $p(\rvx_{t-1} \mid \rvy_{1:t-1}) = \mathcal{N}(\rvx_{t-1} \mid \bm{\mu}_{t-1}, \bm{\Sigma}_{t-1})$ with the predict step,
\begin{align}
    p(\rvx_t | \rvy_{1:t-1}) &= \mathcal{N}(\rvx_t | \bm{\mu}_t^-, \bm{\Sigma}_t^-)\text{, where } %
    \bm{\mu}_t^{-} = \bm{F}_{t-1}\bm{\mu}_{t-1},  %
    \bm{\Sigma}_t^{-} = \bm{F}_{t-1}\bm{\Sigma}_{t-1}\bm{F}_{t-1}^\top + \bm{Q}_{t-1}, %
\end{align}
and then updates that distribution as follows:
\begin{align}
    p(\rvx_t | \rvy_{1:t}) &= \mathcal{N}(\rvx_t \mid \bm{\mu}_t, \bm{\Sigma}_t)\text{, where }  %
    \bm{\mu}_t = \bm{\mu}_t^{-} + \mK_t\rvv_t, %
    \bm{\Sigma}_t = \left[(\bm{\Sigma}_t^{-})^{-1} + \mH_t^\top \mR_t^{-1} \mH_t\right]^{-1},\label{eq:kalman_update_step} %
\end{align}
$\rvv_t = \rvy_t - \mH_t \bm{\mu}_t^-$ is the prediction error, and $\mK_t = \bm{\Sigma}_t^{-}\mH^\top(\mH_{t}\bm{\Sigma}_t^{-}\mH_{t}^\top + \mR_{t})^{-1}$ is known as the Kalman gain. We discuss further details in Appendix~\ref{sec:kalman}.

\subsection{Approximate Bayesian Filtering for Nonlinear Systems}
In real-world systems, guarantees for optimality break down due to inherent nonlinearities and their (at best) approximate Gaussianity. The broader Bayesian filtering community has been motivated by developing methods that can operate in these more realistic settings \citep{sarkka2023bayesian}. Classical solutions include the extended Kalman filter (EKF) and iterated extended Kalman filter (IEKF) that make use of Jacobian matrices to linearize both the dynamics and observation models, and subsequently use the Kalman filtering equations \citep{gelb1974applied}. These are arguably the most popular filtering implementations but only work well in mildly nonlinear systems on short timescales \citep{julier2004unscented}. The unscented Kalman filter (UKF) \citep{julier1995new, julier2004unscented}, like the particle filter (PF) \citep{doucet2001sequential}, is better equipped to handle highly nonlinear systems but both of these methods suffer from the curse of dimensionality and sensitivity to misspecified dynamics. Nonetheless, EKFs, IEKFs, UKFs, and PFs are the most prominent filtering solutions found throughout industry and research, leveraging expensive linearization techniques or particle approximations to treat nonlinearities. We provide complete details for these classical approaches to nonlinear filtering in Appendix~\ref{sec:methods}.

\section{Bayesian Filtering as Optimization}\label{sec:bf_as_opt}
\begin{figure}[t]
\label{explantory}
\begin{center}
\includegraphics[width=\textwidth]{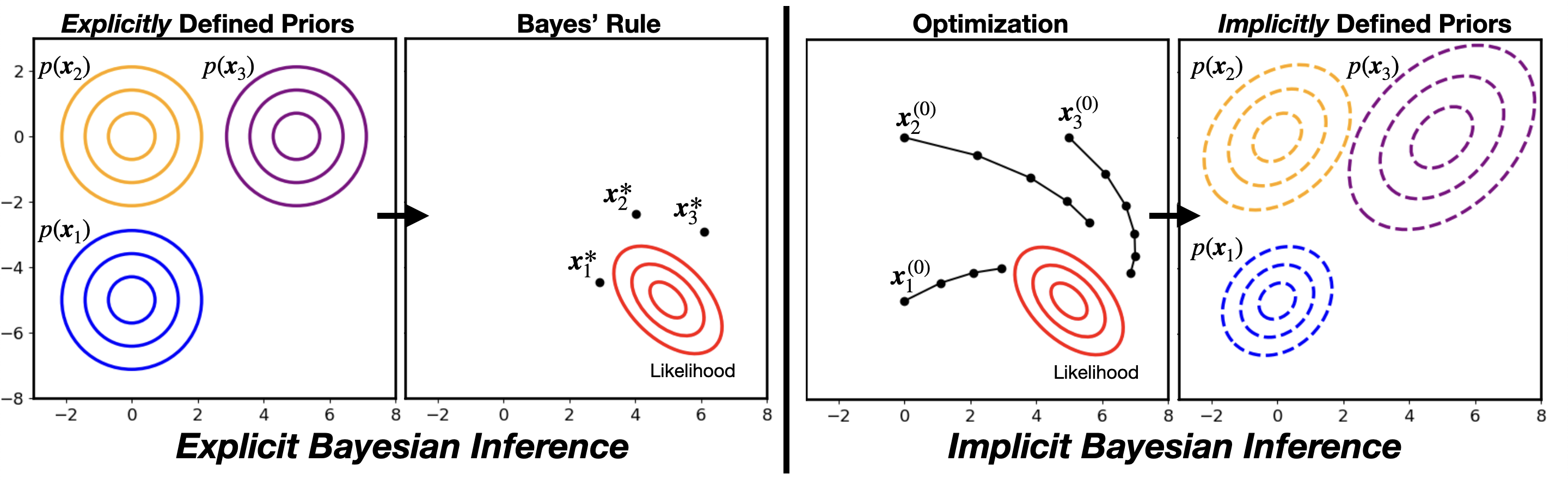}
\end{center}
\caption{\emph{Explicit Bayesian inference} \textbf{(left)} uses Bayes' rule to balance information from the likelihood and explicitly defined priors. $x^*_i$ are MAP estimates of three different posteriors, respective to the priors shown in blue, yellow, and purple. \emph{Implicit Bayesian inference} \textbf{(right)} defines an optimization procedure that reverses the directionality of Bayes' rule by defining a procedure to estimate the posterior mode, which corresponds to an implicitly defined prior. Three trajectories $(x_1^{(0)}, \ldots)$, $(x_2^{(0)}, \ldots)$, $(x_3^{(0)}, \ldots)$ of gradient descent are shown with a fixed learning rate over 3, 4, and 5 steps, respectively. Taking the solutions produced by gradient descent as MAP estimates, we can use the corresponding gradient steps to determine the prior distributions that they implied \citep{santos1996equivalence}.
}
\end{figure}
The update step \cref{eq:update_step} in a Bayesian filter is a straightforward application of Bayes' rule with the predictive distribution $p(\rvx_t | \rvy_{1:t-1})$ playing the role of the prior. In linear-Gaussian systems, this posterior distribution is available in closed-form \cref{eq:kalman_update_step} but one could alternatively achieve an equivalent result via gradient descent due to its connection to regularized least-squares \citep{santos1996equivalence}. In this section, we describe this connection and how an optimization procedure can implicitly define a prior distribution, a kind of dual formulation of Bayes' theorem (see Figure 1). Such a formulation reduces Bayesian inference to point estimates, but this can be more practical for nonlinear systems whose filtering distributions $p(\rvx_t | \rvy_{1:t})$ are not analytically tractable in general.

\subsection{Finding the Posterior Mode via Truncated Gradient Descent}

The mode of the filtering distribution can be expressed as the minimizer of the negative log posterior:
\begin{align}
\bm{\mu}_t = \argmin_{\rvx_t} \bar{\ell}_t(\rvx_t)\text{, where } \bar{\ell}_t(\rvx_t) = -\log p(\rvy_t | \rvx_t) - \log p(\rvx_{t} | \rvy_{1:t-1} ).
\end{align}
In the case of the linear-Gaussian system, the regularized loss function is (up to an additive constant):
\begin{align}
\bar{\ell}_t(\rvx_t) &=  \frac{1}{2} \| \rvy_t - \rmH_t \rvx_t \|^2_{\rmR_t} + \frac{1}{2} \| \rvx_t - \bm{\mu}_t^- \|_{\bm{\Sigma_t^-}}^2. \label{eq:regularized_least_squares},
\end{align}
where $\| \rvz \|^2_{\rmP} \triangleq \rvz^\top \rmP^{-1} \rvz$.
\citet{santos1996equivalence} observed that the minimizer of a regularized least squares problem as in \cref{eq:regularized_least_squares} can be recovered by performing truncated gradient descent on the likelihood term alone. Specifically, let $\bm{\mu}_t^{(0)} = \bm{\mu}_t^-$ and define the recursion
\begin{equation}
    \bm{\mu}_t^{(k+1)} \leftarrow \bm{\mu}_t^{(k)} + \rmM_t  \rmH_t^\top \rmR_t^{-1} (\rvy_t - \rmH_t \bm{\mu}_t^{(k)}), \ \ \text{for} \ \ k = 1, \ldots, K - 1.\label{eq:santos_recurrence}
\end{equation}
Let $\rmB_t$ simultaneously diagonalize $(\bm{\Sigma}^-_t)^{-1}$ and $\rmH_t^\top \rmR_t^{-1} \rmH_t$ such that $\rmB_t^\top (\bm{\Sigma}_t^-)^{-1} \rmB_t = \rmI$ and $\rmB_t^\top \rmH_t^\top \rmR_t^{-1} \rmH_t \rmB_t = \text{diag}(r_1, \ldots, r_n)$. Such a matrix $\rmB_t$ is guaranteed to exist since both $(\bm{\Sigma}^-_t)^{-1}$ and $\rmH_t^\top \rmR_t^{-1} \rmH_t$ are symmetric and $(\bm{\Sigma}^-_t)^{-1}$ is positive definite \citep[Sec.~8.7]{golub2013matrix}. The columns of $\rmB_t$ are the generalized eigenvectors of $\rmH_t^\top \rmR_t^{-1} \rmH_t$ with respect to $(\bm{\Sigma}^-_t)^{-1}$ and $r_1, \ldots, r_n$ are the corresponding eigenvalues. By Theorem 3.1 of \citet{santos1996equivalence}  $\bm{\mu}_t^{(K)}$ defined in \cref{eq:santos_recurrence} is the minimizer of \cref{eq:regularized_least_squares} when $\rmM_t = \rmB_t \text{diag}( \lambda_1, \ldots, \lambda_n) \rmB_t^\top$, where $\lambda_i = (1 / r_i)(1 - (1 + r_i)^{-1/K})$ if $r_i \neq 0$ and $\lambda_i = 1$ otherwise. This suggests an optimization wherein $\bm{\mu}_t$ is computed via $K$ steps of gradient descent on the likelihood, where $\rmM_t$ is the learning rate matrix induced by $\bm{\Sigma}_t^-$.

\subsection{Truncated Gradient Descent as Prior Specification}\label{sec:santos_equivalence}
Instead of computing the learning rate matrix $\rmM_t$ for a given $\bm{\Sigma}_t^-$, consider specifying $\rmM_t$ directly. For example, $\rmM_t = \rho \rmI$ recovers gradient descent on the likelihood with learning rate $\rho$. Let $\rmC_t$ be chosen such that $\rmC_t^\top \rmM_t^{-1} \rmC_t = \rmI$ and $\rmC_t^\top \rmH_t^\top \rmR^{-1}_t \rmH_t \rmC_t = \text{diag}(s_1, \ldots, s_n)$. As for $\rmB_t$ above, such a $\rmC_t$ will exist since both $\rmM_t^{-1}$ and $\rmH_t^\top \rmR^{-1}_t \rmH_t$ are symmetric and $\rmM_t^{-1}$ is positive definite. Then, by Theorem 3.2 of   \citet{santos1996equivalence}, $\bm{\mu}_t^{(K)}$ as defined by \cref{eq:santos_recurrence} is the minimizer of \cref{eq:regularized_least_squares} where $\bm{\Sigma}_t^- = \rmC_t \text{diag}(\sigma_1, \ldots, \sigma_n) \rmC_t^\top$, where $\sigma_i = (1 / s_i) ((1 - s_i)^{-K} - 1)$ if $s_i \neq 0$ and $\sigma_i = 1$ otherwise. Thus, in a linear-Gaussian system, specifying the learning rate matrix $\rmM_t$ implicitly defines an equivalent predictive covariance $\bm{\Sigma}_t^-$ and truncated gradient descent on the likelihood recovers the mode of the filtering distribution.

\subsection{Alternative Interpretation as Variational Inference}
The procedure shown above can be alternatively motivated by positing a variational distribution $q_t(\rvx_t) = \mathcal{N}(\rvx_t \mid \rvm_t, \rmM_t)$ with fixed covariance $\rmM_t$. We show in Appendix~\ref{sec:vi_update} that the truncated optimization of \cref{eq:santos_recurrence} is equivalent to natural gradient ascent on the ELBO. The choice of $\rmM_t$ then specifies the covariance of the variational approximate to the filtering distribution, the learning rate matrix of the optimization, and the effective prior covariance. More generally, \citet{khan2023bayesian} demonstrates that many popular adaptive optimization techniques such as RMSprop and Adam can be interpreted as performing variational inference with different choices of $q_t(\rvx_t)$. In this light, the choice of the optimization algorithm when combined with $k$ steps of optimization implicitly defines a corresponding prior distribution. In the next section, we propose a simple yet effective filtering algorithm based on this interpretation of the update step.

\section{Implicit Maximum a Posteriori Filtering}\label{sec:implicit}

In this section, we state the Bayesian filtering steps of our approach, which (partially) optimizes the likelihood $p(\rvy_t|\rvx_t)$ at every time step. The filtering distribution at each time $t$ is represented by its mean estimate $\bm{\hat{\mu}}_t$.  In addition to the likelihood, we also assume knowledge of a sequence of transition functions $f_t(\rvx_{t-1}) \triangleq \mathbb{E}[\rvx_t | \rvx_{t-1}]$. This is less restrictive than fully specifying the transition distribution, as done in standard Bayesian filtering. The overall algorithm for the Implicit MAP Filter is summarized in \cref{alg:implicit_map_filter}.

\begin{minipage}{0.50\textwidth}
\begin{algorithm}[H]
  \caption{Implicit MAP Filter}
  \label{alg:implicit_map_filter}
  \begin{algorithmic}
  \itemindent=-5pt
    \STATE {\bfseries Input:} Timesteps $T$, initial state estimate $\hat{\bm{\mu}}_0$, sequence of loss functions $\ell_1, \ldots, \ell_T$ where $\exp(-\ell_t(\rvx)) \propto p(\rvy_t \mid \rvx_t = \rvx)$, sequence of transition functions $f_1, \ldots, f_T$ where $f_t(\rvx) = \mathbb{E}[\rvx_t | \rvx_{t-1} = \rvx]$, optimizer $\mathcal{M}$, number of optimization steps $K$. \\
    \STATE {\bfseries Output:} Filtering state estimates $\bm{\hat{\mu}}_1, \ldots, \bm{\hat{\mu}}_T$.
    \vspace{.1cm}
    \FOR{$t=1$ {\bfseries to} $T$}
        \STATE $\bm{\mu}_t^- \leftarrow f_t(\hat{\bm{\mu}}_{t-1})$
        \STATE $\bm{\hat{\mu}}_t \leftarrow \small{\textsc{IMAP-UPDATE}}(\bm{\mu}_t^-, \ell_t, \mathcal{M}, K)$ 
    \ENDFOR
  \end{algorithmic}
\end{algorithm}
\end{minipage}\hfill
\begin{minipage}{0.46\textwidth}
\begin{algorithm}[H]
  \caption{Update step of Implicit MAP Filter: $\small{\textsc{IMAP-UPDATE}}(\bm{\mu}^-, \ell, \mathcal{M}, K)$}
  \label{alg:implicit_map_update}
  \begin{algorithmic}
  \itemindent=-5pt
    \STATE {\bfseries Input:} Initialization $\bm{\mu}^-$, loss function $\ell$, optimizer $\mathcal{M}$, number of optimization steps $K$. \\
    \STATE {\bfseries Output:} State estimate $\bm{\hat{\mu}}$. 
    \vspace{.1cm}
    \STATE $\rvm^{(0)} \leftarrow \bm{\mu}^-$
    \FOR{$k=0$ {\bfseries to} $K-1$}
        \STATE $\rvg^{(k)} \leftarrow \nabla_\rvx \ell(\rvx) \rvert_{\rvx = \rvm^{(k)}}$
        \STATE $\rvm^{(k+1)} \leftarrow \rvm^{(k)} + \mathcal{M}(\rvg^{(k)} ; \rvg^{(0:k-1)})$ 
    \ENDFOR
    \STATE $\bm{\hat{\mu}} \leftarrow \rvm^{(K)}$
  \end{algorithmic}
\end{algorithm}

\end{minipage}
\subsection{Predict Step}\label{sec:predict}
The predictive distribution for the Implicit MAP Filter is obtained by applying the first-order delta method ~\citep{dorfman1938note,verhoef2012who} to the Chapman-Kolmogorov equation \cref{eq:chapman_kolmogorov}:
\begin{align}
    p(\rvx_t \mid \rvy_{1:t-1}) &= \int p(\rvx_t \mid \rvx_{t-1})p(\rvx_{t-1} \mid \rvy_{1:t-1}) d \rvx_{t-1} \approx p(\rvx_t \mid \rvx_{t-1} = \bm{\mu}_{t-1}). 
\end{align}
The mean of the predictive distribution is thus obtained by applying the transition function to the previous state estimate: $\mathbb{E}[\rvx_t | \rvy_{1:t-1}] \approx f_t(\bm{\mu}_{t-1})$. This negates the need for matrix-matrix products as is commonly required for the predict step.

\subsection{Update Step}\label{sec:update}
The update step is simply the application of optimizer $\mathcal{M}$ for $K$ iterations, considering only an initialization $\bm{\mu}_t^-$, observation $\rvy_t$, and loss function $\ell_t$. Here, $\mathcal{M}$ takes as input the history of gradients and outputs a vector to be added to the current state. This formulation captures many popular optimizers, including SGD, RMSProp, and Adam. Our algorithm for the update step of the Implicit MAP Filter is provided in \cref{alg:implicit_map_update}.

When choosing $\mathcal{M}$ and $K$, the goal is to specify an inductive bias that corresponds to the true filtering equations for the system. Choices such as increasing $K$ and the learning rate increase the Kalman gain \citep{freitas2000hierarchical}. The nature of each optimization step, as well as $K$, defines the shape and entropy of the filtering distribution assumed \citep{duvenaud2016early}. Changing the initialization $\bm{\mu}_t^-$ given by the predict step modifies the location of the \emph{implicitly} defined prior distribution within the latent space. This concert of initialization $\bm{\mu}_t^-$, optimizer $\mathcal{M}$, and number of iterations $K$ produces a state estimate $\bm{\hat{\mu}}_t$ that maximizes the posterior with respect to an \emph{implicitly} defined prior. The aim when specifying these quantities is to define the \emph{correct} posterior.

\section{Experimental Evaluation}\label{sec:experiments}

The goal of our experiments is to assess whether the most popular adaptive optimizers can be used to design effective, robust, and scalable Bayesian filters. We begin with a simple low-dimensional system with nonlinear dynamics to establish that our approach is competitive with standard filters. Then, we turn to a more challenging system that has been used as a benchmark for these algorithms, demonstrating our advantages. Finally, we show how Implicit MAP Filtering can be effective in high-dimensional settings, such as adapting the weight space of a convolutional neural network.

\subsection{Toy nonlinear System}\label{sec:toy}

The equivalence between Kalman filtering and $K$ steps of gradient descent elucidated in Section~\ref{sec:santos_equivalence} suggests that the proposed framework should produce reliable estimates in \emph{approximately} linear-Gaussian systems. To better understand filtering amidst nonlinearites, we first consider the one-dimensional example used in \citet{doucet2001sequential,gordon1993novel,kitagawa1996monte}, originally proposed by \citet{netto1978new}, which admits highly nonlinear periodic behavior (see Figure 2):
\begin{align}
    \rvx_{t} = \frac{1}{2}\rvx_{t-1} + 25\frac{\rvx_{t-1}}{1 + \rvx_{t-1}^2} + 8\cos{(1.2t\Delta t)} + \bm{q}_{t-1}, \quad \quad \rvy_{t} = \frac{\rvx_t^2}{20} + \bm{r}_{t},
\end{align}
where $\rvx_{0} \sim \mathcal{N}(0, 1)$, $\bm{q}_{t-1} \sim \mathcal{N}(0, \bm{Q}_{t-1})$, $\bm{r}_{t} \sim \mathcal{N}(0, \bm{R}_{t})$, and $\Delta t = 0.1$. We simulate ground truth trajectories and their measurements for 200 time steps and evaluate performance using the root mean square errors (RMSEs) with respect to the simulated true trajectories for every combination of $\bm{Q}_{t-1} \in \{1, 3, 5\}$ and $\bm{R}_{t} \in \{1, 2, 3\}$. The means and $95\%$ confidence intervals of the RMSEs are computed from 100 independent Monte Carlo (MC) simulations. For each optimizer, we report the best hyperparameters found by a grid search using 5 separate MC simulations (see Appendix \ref{sec:grid}). 

\begin{figure}[t]
\label{system1figure}
\begin{center}\includegraphics[width=\textwidth]{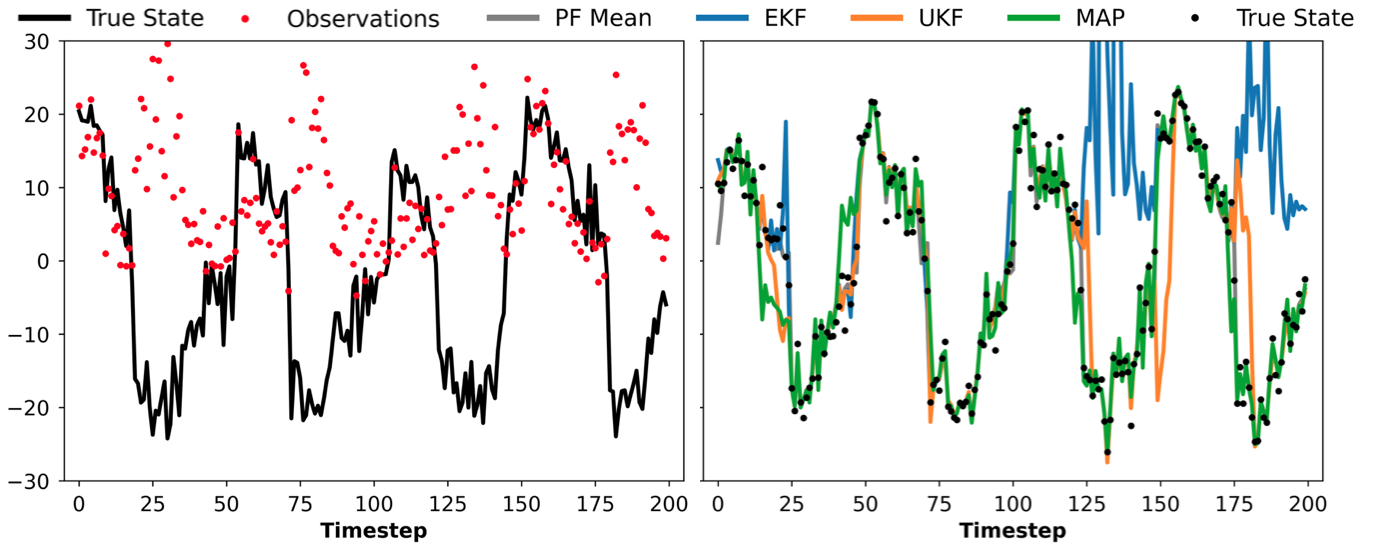}
\end{center}
\caption{Toy nonlinear system ($\bm{Q}_{t-1} = 3, \bm{R}_{t} = 2)$ for a single random trajectory \textbf{(left)} and the respective filter estimates \textbf{(right)}. Implicit MAP Filter (MAP) shown uses the Adam optimizer with 50 steps at learning rate $\eta = 0.1$ and exponential decay rates $\beta_1, \beta_2 = 0.1$.}
\end{figure}

Table \ref{experiment_1_table} shows the results for $\bm{R}_t = 2$ and $\bm{Q}_{t-1} = 1, 3, 5$. There exist RMSprop and Adam hyperparameters that are competitive with the unscented Kalman filter (UKF) under low-to-medium process noise and better under high process noise. Both the extended Kalman filter (EKF) and iterated extended Kalman filter (IEKF) have the tendency to diverge in this system. The particle filter (PF) performs the best universally, since 1000 particles can capture uni-modal and bi-modal filtering distributions in one dimension quite well. Nonetheless, our Implicit MAP Filter tends to produce estimates that match the modes of the particle filter (see Appendix~\ref{sec:extended1}) only struggling on some parts of the sequence with rapid changes in the state. We describe the implementation details for these baseline methods in Appendix~\ref{sec:methods} and show similar results for $\bm{R}_t = 1, 3$ in Appendix~\ref{sec:extended1}.

The Implicit MAP Filter does not explicitly use $\bm{Q}_{t-1}$ or $\bm{R}_t$ but internalizes both of these terms by the choice of optimizer. In fact, our Implicit MAP Filter can always arbitrarily set $\bm{R}_t = \bm{I}$, reducing the objective to the mean squared error loss (see Appendix \ref{sec:justification} for justification). Compared to methods that similarly use gradient information (EKF, IEKF), our Implicit MAP Filters with Adam and RMSprop see a significant performance boost despite the challenge of a non-convex likelihood. We can attribute this boost to the use of momentum (see Appendix \ref{sec:extended1}). In contrast to methods that use sigma points (UKF) or particles (PF), we see competitive performance while avoiding covariance calculations or maintaining many copies of the state.

\begin{table}[t]
\caption{RMSEs on the toy nonlinear system ($\bm{R}_t = 2$). Results show the average RMSE over 100 MC simulations with 95\% confidence intervals.}
\label{experiment_1_table}
\centering
\renewcommand{\arraystretch}{1.0} %
\begin{tabular}{lccc}
\toprule
\textbf{Method} & \textbf{RMSE ($\bm{Q}_{t-1} = 1$)} & \textbf{RMSE ($\bm{Q}_{t-1} = 3$)} & \textbf{RMSE ($\bm{Q}_{t-1} = 5$)} \\
\midrule
EKF & $24.619 \pm 3.988$ & $34.909 \pm 2.732$ & $45.646 \pm 5.629$ \\
IEKF ($K = 5$) & $9.277 \pm 0.652$ & $15.321 \pm 0.493$ & $17.898 \pm 0.559$ \\
\midrule
IMAP (Adadelta)* & $32.008 \pm 8.650$ & $23.152 \pm 3.973$ & $26.462 \pm 5.058$ \\
IMAP (Gradient Descent)* & $5.589 \pm 0.219$ & $7.966 \pm 0.180$ & $10.130 \pm 0.196$ \\
IMAP (Adagrad)* & $5.569 \pm 0.215$ & $6.549 \pm 0.223$ & $9.264 \pm 0.227$ \\
IMAP (RMSprop)* & $5.304 \pm 0.256$ & $6.000 \pm 0.227$ & $8.527 \pm 0.441$ \\
IMAP (Adam)* & $5.699 \pm 0.190$ & $5.842 \pm 0.231$ & $7.964 \pm 0.334$ \\
\midrule
UKF & $4.550 \pm 0.242$ & $5.762 \pm 0.270$ & $9.555 \pm 0.400$ \\
\midrule
PF ($n = 1000$) & $\bm{1.565 \pm 0.047}$ & $\bm{2.800 \pm 0.108}$ & $\bm{4.518 \pm 0.160}$ \\
\bottomrule
\end{tabular}
*Methods where the reported hyperparameters were found via grid search (see Appendix \ref{sec:grid}).
\end{table}

\subsection{Stochastic Lorenz Attractor}\label{sec:lorenz}
In the previous section, we show that our approach is competitive with a particle filter (PF) and the unscented Kalman filter (UKF) in a system that is relatively easy for these methods. Now, we evaluate a mildly nonlinear chaotic system that is more complicated because the stochastic differential equation (SDE) is unreliable for accurately predicting the next state \citep{evensen2009data}.

Consider a stochastic Lorenz '63 system \citep{lorenz1963deterministic, li2020scalable, zhao2021non},
\begin{align}
    \text{d} \begin{pmatrix} x_{1,t} \\ x_{2,t} \\ x_{3,t} \end{pmatrix} = \begin{pmatrix}
\sigma (x_{2,t} - x_{1,t}) \\ x_{1,t} (\rho - x_{3,t}) - x_{2,t} \\ x_{1,t} \cdot x_{2,t} - \beta x_{3,t}
\end{pmatrix} \text{d}t + \alpha \text{d}W_t, \quad \quad \rvy_t = \begin{pmatrix} x_{1,t} \\ x_{2,t} \\ x_{3,t} \end{pmatrix} + \bm{r}_t,
\end{align}
where $\rvx_{0} \sim \mathcal{N}(\bm{\mu}_0, \bm{I}_3)$ for $\bm{\mu}_0 = (10, 10, 10)$, $\bm{r}_{t} \sim \mathcal{N}(\bm{0}, 2 \cdot \bm{I}_3)$, $\sigma = 10$, $\rho = 28$, $\beta = \frac{8}{3}$, $\alpha = 10$, and $W_t$ is a three-dimensional Wiener process with unit spectral density. Similar to \citet{zhao2021non}, we simulate numerical ground truth trajectories and their measurements uniformly for 200 time steps with d$t = 0.02$ using 10000 Euler-Maruyama integration steps between each measurement. We compute RMSEs and $95\%$ confidence intervals from 100 independent MC simulations and report the best hyperparameters found by a separate grid search (see Appendix \ref{sec:grid}).

We report three experimental configurations where we only vary the method used for numerical integration: 4th order Runge-Kutta (RK4) method, Euler's method (Euler), and no integration, assuming a Gaussian random walk (GRW). These methods are presented in decreasing order of accuracy with respect to the aforementioned Euler-Maruyama integration steps. The objective is to assess robustness to dynamics misspecification, which is an unavoidable reality wherever numerical integration is required. Unlike the toy nonlinear system, where the true process noise covariance is known exactly, numerical integration introduces additional unmodeled noise to the process. This presents a realistic challenge that can significantly distort estimates \citep{huang2017novel, mehra1972approaches}.

Table \ref{experiment_2_table} shows the results of these experiments and the number of values stored to describe the state ($N$). For fair comparison, we show the EKF and UKF with $\bm{Q}_{t-1}$ optimized by grid search (see Appendix \ref{sec:grid1}) in addition to the EKF and UKF with the $\bm{Q}_{t-1}$ specified by the system. Performing such a search for the PF is impractical due to the cost of numerically integrating each particle. The Implicit MAP Filter with gradient descent and the EKF with optimized $\bm{Q}_{t-1}$ show the best performance. Both are relatively robust to worsening estimates of the SDE. The EKF with optimized $\bm{Q}_{t-1}$ is close to optimal due to the convexity of this system.

The performance of the Implicit MAP Filter and the optimized EKF are statistically equivalent but with several key differences. Our approach only computes a point estimate over a minimal state space representation, which decreases space complexity and avoids expensive matrix-matrix computations. Classical filters such as the EKF are difficult to tune \citep{chen2018weak} whereas simple adaptive optimizers, like Adam, are fixed to 3 or 4 hyperparameters regardless of the state space and are faster to evaluate, making search quick and effective. The EKF is not well suited for nonconvexities, as shown in Section~\ref{sec:toy}, whereas the Implicit MAP Filter overcomes this pitfall by its choice of optimizer.

Accommodating worsening dynamics estimates is straightforward by either increasing the number of steps or learning rate, each of which corresponds to a weaker prior. In Table \ref{experiment_2_table}, our Implicit MAP Filter with gradient descent uses 3 gradient steps with learning rate $\eta = 0.05$ with RK4, 3 steps with $\eta = 0.1$ for the Euler case, and 10 steps with $\eta = 0.1$ for the GRW case. However, the same optimizer settings generally show robustness from RK4 to Euler (see Appendix \ref{sec:extended2}). When visualizing trajectories (see Appendix \ref{sec:extended2}), our approach does a remarkable job of fitting this system despite every initialization producing a vastly different filtering problem. We report additional results for this system using $\alpha = 1, 5, 20$ in Appendix \ref{sec:extended2}.

\begin{table}[t]
\caption{RMSEs on the stochastic Lorenz attractor ($\alpha = 10$). Results show the average RMSE over 100 MC simulations with 95\% confidence intervals. RK4 indicates 4th order Runge-Kutta method. Euler indicates Euler's method. GRW indicates a Gaussian random walk. $N$ is the number of values stored to describe the state space ($D = 3$).}
\label{experiment_2_table}
\centering
\renewcommand{\arraystretch}{1.0} %
\begin{tabular}{lcccc}
\toprule
\textbf{Method} & \textbf{RMSE (RK4)} & \textbf{RMSE (Euler)} & \textbf{RMSE (GRW)} & $\bm{N}$ \\
\midrule
EKF (original $\bm{Q}_{t-1}$) & $0.890 \pm 0.094$ & $2.308 \pm 0.155$ & $3.057 \pm 0.037$ & $D + D^2$ \\
EKF (optimized $\bm{Q}_{t-1}$)* & $\bm{0.692 \pm 0.014}$ & $\bm{0.952 \pm 0.011}$ & $\bm{1.561 \pm 0.010}$ & $D + D^2$ \\
IEKF ($K = 5$) & $0.890 \pm 0.094$ & $2.308 \pm 0.158$ & $3.057 \pm 0.038$ & $D + D^2$ \\
\midrule
IMAP (Adadelta)* & $6.034 \pm 0.634$ & $7.649 \pm 0.254$ & $10.754 \pm 0.279$ & $\bm{D}$ \\
IMAP (Adam)* & $0.982 \pm 0.135$ & $1.153 \pm 0.014$ & $1.885 \pm 0.010$ & $\bm{D}$ \\
IMAP (Adagrad)* & $0.907 \pm 0.144$ & $1.096 \pm 0.017$ & $1.765 \pm 0.013$ & $\bm{D}$ \\
IMAP (RMSprop)* & $0.881 \pm 0.114$ & $1.081 \pm 0.015$ & $1.757 \pm 0.015$ & $\bm{D}$ \\
IMAP (Gradient Descent)* & $\bm{0.701 \pm 0.018}$ & $\bm{0.960 \pm 0.012}$ & $\bm{1.561 \pm 0.010}$ & $\bm{D}$ \\
\midrule
UKF (original $\bm{Q}_{t-1}$) & $2.742 \pm 0.059$ & $2.856 \pm 0.066$ & $5.628 \pm 0.067$ & $D + D^2$ \\
UKF (optimized $\bm{Q}_{t-1}$)* & $1.402 \pm 0.010$ & $1.417 \pm 0.011$ & $1.736 \pm 0.013$ & $D + D^2$ \\
\midrule
PF ($n = 1000$) & $1.568 \pm 0.027$ & $1.725 \pm 0.031$ & $14.346 \pm 0.365$ & $nD$ \\
\bottomrule
\end{tabular}
*Methods where the reported hyperparameters were found via grid search (see Appendix \ref{sec:grid}).
\end{table}

\subsection{Yearbook}\label{sec:yearbook}

Suppose we have a pre-trained neural network that performs inference in a time-varying environment. This could be a recommender system that must evolve with changing seasons or a facial recognition system that must be robust to temporal shift. Let us assume that data curation is expensive, so we are only given a small number of examples at every time step for adaptation. We can formalize this as the following state space model \citep{haykin2004kalman}:
\begin{align}
    \rvw_{t} = f(\rvx_{t-1}) + \bm{q}_{t-1}, \quad \quad
    \rvy_{t} = h(\rvw_{t}, \bm{X}_t) + \bm{r}_{t}
\end{align}
where $\rvw_{t}$ is the network's \emph{ideal} weight parameterization, $\bm{q}_{t-1}$ is some noise vector that corrupts estimates from the transition function $f$, $\rvy_{t}$ is the network's desired response vector after applying the transformation $h$ parameterized by $\rvw_{t}$ with feature matrix $\bm{X}_{t}$, and $\bm{r}_{t}$ is the measurement noise. 

In this final experiment, we design a time-varying prediction task as described above using the Yearbook dataset~\citep{ginosar2015century}, which contains 37,921 frontal-facing American high school yearbook photos that capture changes in fashion and population demographics from 1905 - 2013. We use the data from 1905-1930 to pre-train a 4 layer convolutional neural network (CNN) to perform a binary gender classification task. From 1931-2010, we adapt this weight initialization at each year sequentially using five simple approaches: static weights, direct fit, variational Kalman filtering (VKF), particle filtering (PF), and our Implicit MAP filtering. 32 randomly chosen training examples are available at every time step and a held-out test set of roughly 100 images per time step is used validate the approaches (see Appendix \ref{sec:extended3} for complete details).

This system is complicated by (1) a state space of 28,193 parameters, (2) unknown weight-space dynamics equations, and (3) unknown process noise and measurement noise covariances. To understand our capacity for modeling such a system given a VKF, PF, and Implicit MAP Filter, we test 5 hyperparameter configurations of each method (see Appendix \ref{sec:extended3}). We report classification accuracies for all five methods in Table \ref{yearbook_table}, grouped into two 40 year blocks. Static weights yields the worst performance due to the problem being fundamentally non-stationary. On the other extreme, attempting to exactly fit the data at each time step severely overfits, which proves to be better but still suboptimal. The VKF, PF, and Implicit MAP Filter attempt to internalize the system's true state space equations via a small grid search, but this is clearly difficult to do exactly in such a state space. Our implicit approach is less sensitive to this inevitable misspecification. The Implicit MAP Filter with 50 steps of Adam optimization not only showed the best performance, but also highly desirable space and time complexity. In cases where we do not start with a pre-trained network, we can simply fit the first timestep to convergence, which is equivalent to an uninformative prior on the initial state.

\begin{table}[t]
\caption{Classification accuracy with $95\%$ confidence intervals over two 40 year blocks in the yearbook dataset.}
\label{yearbook_table}
\centering
\renewcommand{\arraystretch}{1.0} %
\begin{tabular}{lcc}
\toprule
\textbf{Method} & \textbf{\% Correct (Validation Years)} & \textbf{\% Correct (Test Years)} \\
\midrule
Static Weights & $82.416 \pm 2.270$ & $60.897 \pm 1.615$\\
Particle Filter & $86.430 \pm 1.790$ & $66.715 \pm 2.390$\\
Variational Kalman Filter & $93.087 \pm 1.038$ & $79.967 \pm 2.204$\\
Direct Fit & $94.416 \pm 0.924$ & $80.449 \pm 1.845$ \\
Implicit MAP Filter & $\bm{94.973 \pm 0.837}$ & $\bm{84.747 \pm 2.030}$ \\
\bottomrule
\end{tabular}
\end{table}

\section{Related Work}\label{sec:related}
Bayesian filtering, optimization, and neural networks share a long history that dates back to early attempts of using the extended Kalman filter (EKF) algorithm to train multilayer perceptrons \citep{singhal1988training}. Since then, Kalman filter theory for the training and use of neural networks saw rich development \citep{haykin2004kalman}, but fell out of fashion as gradient descent proved to be the more efficient and scalable option for the training of deep neural networks \citep{bengio2017deep}. %
\citet{ruck1992comparative} realized a connection between optimization and Bayesian filtering, showing that the standard backpropagation algorithm can be seen as a degenerate form of the EKF. Several works have since connected the EKF and the Gauss-Newton method \citep{bertsekas1996incremental, bell1993iterated} and showed that the EKF can be seen as a single iteration of Newton’s method for a specific quadratic form \citep{humpherys2012fresh}. %
\citet{ollivier2018online,ollivier2019extended} proved equivalence between online natural gradient descent and the EKF when the dynamics are stationary or the process noise is proportional to the posterior covariance over states. \citet{aitchison2020bayesian} used the Bayesian filtering formalism to derive adaptive optimizers like Adam and RMSprop. \citet{freitas2000hierarchical} showed that the process noise and measurement noise can be viewed as adaptive per-parameter learning rates. In the present work, we establish similar connections between optimization and Bayesian filtering but focus on optimization as a tool for Bayesian filtering rather than Bayesian filtering as a tool for optimization.

Several works have similarly used popular optimization methods to address challenges with classical Bayesian filters. \citet{auvinen2010variational} and \citet{bardsley2013krylov} used limited memory Broyden–Fletcher–Goldfarb–Shanno (LBFGS) and conjugate gradients, respectively, to derive variational Kalman filters with low storage covariance, and inverse covariance, approximations. These methods scale up classical filters, like low-rank extended Kalman filters \cite{chang2023low}, rather than providing a general filtering approach that is painless to implement. \citet{chen2003bayesian}  describes particle filters that move every particle down their respective gradient prior to sampling from the proposal distribution. 
This is reminiscent of our approach in that it uses gradients, but differs in that it maintains particles, performs sampling, and again falls within the class of explicit filters.

\section{Conclusion}\label{sec:conclusion}

We have shown that Bayesian filtering can be considered as optimization over a time-varying objective and that such a perspective opens the door to effective, robust, and scalable filters built from adaptive optimizers. This framework, however, comes at the cost of interoperability and uncertainty estimates, which limits use in risk-sensitive environments or situations where the filtering equations themselves are used for analysis. Nonetheless, our proposed Implicit MAP Filtering approach is an attractive option for the performance-driven practitioner.

\subsubsection*{Reproducibility Statement}
All results and figures reported can be reproduced with the code made available at: \href{https://github.com/gianlucabencomo/implicitMAP}{github.com/gianlucabencomo/implicitMAP}.

\subsubsection*{Acknowledgments}
We thank R. Thomas McCoy, Jianqiao Zhu, Logan Nelson, and Ryan Adams for helpful discussions. JCS was supported by the Schmidt DataX Fund at Princeton University made possible through a major gift from the Schmidt Futures Foundation. GMB, JCS, and TLG were supported by grant N00014-23-1-2510 from the Office of Naval Research.

\bibliography{iclr2024_conference}
\bibliographystyle{iclr2024_conference}

\newpage

\appendix
\section{Appendix}\label{sec:appendix}

\subsection{Baseline Bayesian Filter Methods and Implementation Details}\label{sec:methods}

In this section we define the Kalman filter (KF), Extended Kalman filter (EKF), iterated extended Kalman filter (IEKF), unscented Kalman filter (UKF), particle filter (PF), and variational Kalman filter (VKF) as implemented in the experimental section of this paper. We closely follow the definitions and notation of \citet{sarkka2023bayesian} and refer the reader to that text for proofs and a more extensive treatment of these topics. 

We use this section to state design choices made for each baseline filter when it comes to the experiments reported in the main section.

\subsubsection{Kalman Filter}\label{sec:kalman}
Assume the following state space model:
\begin{align*}
    \rvx_t &= \bm{F}_{t-1}\rvx_{t-1} + \bm{q}_{t-1} \\
    \rvy_t &= \bm{H}_t\rvx_t + \bm{r}_t
\end{align*}
where $\rvx_t \in \mathbb{R}^n$ is the state, $\rvy_t \in \mathbb{R}^m$ is the measurement, $\bm{F}_{t-1} \in \mathbb{R}^{n \times n}$ is the transition matrix, $\bm{H}_{t} \in \mathbb{R}^{m \times n}$ is the measurement model matrix, $\bm{q}_{t-1} \sim \mathcal{N}(\bm{0}, \bm{Q}_{t-1})$ is the process noise and $\bm{r}_{t} \sim \mathcal{N}(\bm{0}, \bm{R}_{t})$ is the measurement noise.

The Kalman filter \citep{kalman1960new} defines the predictive distribution, filtering distribution, and marginal likelihood at time step $t$ as
\begin{align*}
    p(\rvx_t|\rvy_{1:t-1}) &= \mathcal{N}(\rvx_t|\bm{\mu}_t^{-}, \bm{\Sigma}_t^{-}), \\
        p(\rvx_t|\rvy_{1:t}) &= \mathcal{N}(\rvx_t|\bm{\mu}_t, \bm{\Sigma}_t), \\
        p(\rvy_t|\rvy_{1:t-1}) &= \mathcal{N}(\rvx_t|\bm{H}_t\bm{\mu}_t^{-}, \bm{S}_t),
\end{align*}
with predict step:
\begin{align*}
    \bm{\mu}_t^{-} &= \bm{F}_{t-1}\bm{\mu}_{t-1}, \\
    \bm{\Sigma}_t^{-} &= \bm{F}_{t-1}\bm{\Sigma}_{t-1}\bm{F}_{t-1}^\top + \bm{Q}_{t-1},
\end{align*}
and update step:
\begin{align*}
    \bm{v}_t &= \rvy_t - \bm{H}_{t}\bm{\mu}_t^{-}, \\
            \bm{S}_t, &= \bm{H}_{t}\bm{\Sigma}_t^{-}\bm{H}_{t}^\top + \bm{R}_{t}, \\
            \bm{K}_t &= \bm{\Sigma}_t^{-}\bm{H}_{t}^\top\bm{S}_t^{-1}, \\
            \bm{\mu}_t &= \bm{\mu}_t^{-} + \bm{K}_t\bm{v}_t, \\
            \bm{\Sigma}_t &= \bm{\Sigma}_t^{-} - \bm{K}_t\bm{S}_t\bm{K}_t^\top.
\end{align*}
Recursion begins from some $\rvx_0 \sim \mathcal{N}(\bm{\mu}_0, \bm{\Sigma}_0)$. In the stochastic Lorenz Attractor, the Gaussian random walk (GRW) case is equivalent to a Kalman filter where $\bm{F}_{t-1} = \bm{I}_3$ and $\bm{H}_{t} = \bm{I}_3$ at every time step $t$. We do not see the linear-Gaussian Kalman filter in any other experiments.

\subsubsection{Extended Kalman Filter}\label{sec:ekf}

In the extended Kalman filter (EKF), we use first-order Taylor approximations of the nonlinear transition function $f$ and measurement model $h$ where appropriate. We denote the Jacobians of these functions as $\bm{F}_{\rvx}(\cdot)$ and $\bm{H}_{\rvx}(\cdot)$. The predict step is now
\begin{align*}
    \bm{\mu}_t^{-} &= f(\bm{\mu}_{t-1}), \\
    \bm{\Sigma}_t^{-} &= \bm{F}_{\rvx}(\bm{\mu}_{t-1})\bm{\Sigma}_{t-1}\bm{F}_{\rvx}(\bm{\mu}_{t-1})^\top + \bm{Q}_{t-1},
\end{align*}
and the update step is
\begin{align*}
    \bm{v}_t &= \rvy_t - h(\bm{\mu}_t^{-}), \\
            \bm{S}_t, &= \bm{H}_{\rvx}(\bm{\mu}_t^{-})\bm{\Sigma}_t^{-}\bm{H}_{\rvx}(\bm{\mu}_t^{-})^\top + \bm{R}_{t}, \\
            \bm{K}_t &= \bm{\Sigma}_t^{-}\bm{H}_{\rvx}(\bm{\mu}_t^{-})^\top\bm{S}_t^{-1}, \\
            \bm{\mu}_t &= \bm{\mu}_t^{-} + \bm{K}_t\bm{v}_t, \\
            \bm{\Sigma}_t &= \bm{\Sigma}_t^{-} - \bm{K}_t\bm{S}_t\bm{K}_t^\top,
\end{align*}
with recursion starting from some $\rvx_0 \sim \mathcal{N}(\bm{\mu}_0, \bm{\Sigma}_0)$ as before. 

\subsubsection{Iterated Extended Kalman Filter}\label{sec:iekf}

The iterated extended Kalman filter (IEKF) \citep{gelb1974applied} shares an identical predict step with the EKF. The update step is redefined as the following iterative procedure, starting from $\rvx^{(0)}_t = \bm{\mu}^{-}_t$ from the predict step:
\begin{itemize}
    \item For $i = 1, 2, \ldots K$:
    \begin{align*}
        \bm{v}_t^{(i)} &= \rvy_t - h(\rvx^{(i-1)}_t) - \bm{H}_{\rvx}(\rvx^{(i-1)}_t)(\bm{\mu}^{-} - \rvx^{(i-1)}_t), \\
            \bm{S}_t^{(i)}, &= \bm{H}_{\rvx}(\rvx^{(i-1)}_t)\bm{\Sigma}_t^{-}\bm{H}_{\rvx}(\rvx^{(i-1)}_t)^\top + \bm{R}_{t}, \\
            \bm{K}_t^{(i)} &= \bm{\Sigma}_t^{-}\bm{H}_{\rvx}(\rvx^{(i-1)}_t)^\top\left[\bm{S}_t^{(i)}\right]^{-1}, \\
            \rvx^{(i)}_t &= \bm{\mu}_t^{-} + \bm{K}_t^{(i)}\bm{v}_t^{(i)}.
    \end{align*}
    \item Set $\bm{\mu}_t = \rvx^{(K)}_t$.
    \item Set $\bm{\Sigma}_t = \bm{\Sigma}_t^{-} - \bm{K}_t^{(K)}\bm{S}_t^{(K)}\left[\bm{K}_t^{(K)}\right]^\top$.
\end{itemize}
For both the stochastic Lorenz Attractor and the toy nonlinear system, we ran IEKFs for iterations $K \in \{1, 3, 5, 10, 25, 50, 100\}$. We consistently report the IEKF that showed the highest average RMSE over 100 Monte Carlo experiments.

\subsubsection{Unscented Kalman Filter}\label{sec:ukf}

The predict step for the Unscented Kalman filter (UKF) is as follows:
\begin{itemize}
    \item Form $2n + 1$ sigma points where 
    \begin{align*}
        \chi_{t-1}^{(0)} &= \bm{\mu}_{t-1}, \\
        \chi_{t-1}^{(i)} &= \bm{\mu}_{t-1} + \sqrt{n + \lambda} \left[\sqrt{\bm{\Sigma}_{t-1}}\right]_{i}, \\
        \chi_{t-1}^{(i + n)} &= \bm{\mu}_{t-1} - \sqrt{n + \lambda} \left[\sqrt{\bm{\Sigma}_{t-1}}\right]_{i},
    \end{align*}
    for $i = 1, \ldots, n$ where $n$ is the dimension of the state space and $\lambda$ is a tuneable parameter defined below.
    \item Apply the dynamics model to the sigma points:
    \begin{align*}
        \hat{\chi}_{t}^{(i)} = f(\chi_{t-1}^{(i)})
    \end{align*}
    for $i = 0, \ldots, 2n$.
    \item Compute the predict step mean and covariance:
    \begin{align*}
    \bm{\mu}_t^{-} &= \sum_{i=0}^{2n} w_i^{(m)} \hat{\chi}_{t}^{(i)}, \\
    \bm{\Sigma}_t^{-} &= \sum_{i=0}^{2n} w_i^{(c)} \left(\hat{\chi}_{t}^{(i)} - \bm{\mu}_t^{-}\right)\left(\hat{\chi}_{t}^{(i)} - \bm{\mu}_t^{-}\right)^\top + \bm{Q}_{t-1},
\end{align*}
where
\begin{align*}
    w_0^{(m)} &= \frac{\lambda}{n + \lambda},\\
    w_0^{(c)} &=  \frac{\lambda}{n + \lambda} + (1 - \alpha^2 + \beta),\\
    w_i^{(m)} &= \frac{\lambda}{2(n + \lambda)},\\
    w_i^{(c)} &= \frac{\lambda}{2(n + \lambda)},\\
\end{align*}
where $\alpha, \beta$ are tuneable parameters.
\end{itemize}
The update step:
\begin{itemize}
    \item Form $2n + 1$ sigma points as before except replace $\bm{\mu}_{t-1}$ and $\bm{\Sigma}_{t-1}$ with $\bm{\mu}_t^{-}$ and $\bm{\Sigma}_t^{-}$ from the predict step. Denote these sigma points as $\hat{\chi}_{t}^{-(i)}$ for $i = 0, \ldots, 2n.$
    \item Apply the measurment model to the sigma points:
    \begin{align*}
        \hat{\mathcal{Y}}_{t}^{(i)} = h(\chi_{t-1}^{(i)}),
    \end{align*}
    for $i = 0, \ldots, 2n$.
    \item Compute:
    \begin{align*}
        \bm{m}_t &= \sum_{i=0}^{2n} w_i^{(m)} \hat{\mathcal{Y}}_{t}^{(i)},  \\
        \bm{S}_t &= \sum_{i=0}^{2n} w_i^{(c)} \left(\hat{\mathcal{Y}}_{t}^{(i)} - \bm{m}_t\right)\left(\hat{\mathcal{Y}}_{t}^{(i)} - \bm{m}_t\right)^\top + \bm{R}_{t},\\
        \bm{C}_t &= \sum_{i=0}^{2n} w_i^{(c)} \left(\hat{\chi}_{t}^{-(i)} - \bm{\mu}_t^{-}\right)\left(\hat{\mathcal{Y}}_{t}^{(i)} - \bm{m}_t\right)^\top. \\
    \end{align*}
    \item Compute the Kalman gain and perform the update:
    \begin{align*}
        \bm{K}_t &= \bm{C}_t\bm{S}_t^{-1}, \\
        \bm{\mu}_t &= \bm{\mu}_t^{-} + \bm{K}_t\left(\rvy_t - \bm{m}_t\right), \\
        \bm{\Sigma}_t &= \bm{\Sigma}_t^{-} - \bm{K}_t\bm{S}_t\bm{K}_t^\top.
    \end{align*}
\end{itemize}
For both the stochastic Lorenz Attractor and the toy nonlinear system, we set $\alpha = 1$, $\beta = 3 - n$, $\lambda = \alpha^2 \cdot (n + \beta) - n$.

\subsubsection{Particle Filter}\label{sec:pf}

We ran a Bootstrap filter (BF) with 1000 particles for every experiment reported in the main section. The BF algorithm at every time step is as follows:
\begin{itemize}
    \item Sample 
    \begin{align*}
        \rvx_t^{(i)} \sim p(\rvx_t|\rvx_{t-1}^{(i)})
    \end{align*}
    for every $i$th particle.
    \item Compute weights
    \begin{align*}
        w_t^{(i)} \propto p(\rvy_t|\rvx_t^{(i)})
    \end{align*}
    for every $i$th particle and normalize.
    \item Perform resampling.
\end{itemize}
We did multinomial resampling at every time step $t$. For the yearbook dataset, since the transition distribution was unknown, we used the following transition distribution:
\begin{align}
    p(\rvx_t|\rvx_{t-1}^{(i)}) = \mathcal{N}(p(\rvx_t|\rvx_{t-1}^{(i)}, \sigma^2\bm{I})
\end{align}
where $\sigma^2 \in \{0.1, 0.05, 0.01, 0.005, 0.001\}$. Selecting $\sigma^2$ requires a grid search, which is described in Appendix \ref{sec:grid}.

\subsubsection{Variational Kalman Filter}\label{sec:vkf}

A variational Kalman filter (VKF) was used in the Yearbook experiments due to the storage complexity of the EKF and UKF. Similar to the Implicit MAP Filter, the VKF we implemented propagates a point-mass forward in time but differs in that it models the prior uncertainty \emph{explicitly} rather than \emph{implicitly}.

Assume that the initial distribution is $p(\rvw_0) = \mathcal{N}(\rvw_0|\bm{0}, \sigma^2_0 \bm{I})$, the transition distribution is $p(\rvw_t|\rvw_{t-1}) = \mathcal{N}(\rvw_t|\rvw_{t-1}, \sigma^2\bm{I})$, and the likelihood is $p(\rvy_t|\bm{X}_t) = \prod_{i=1}^{n_t}p_{\rvw_t}(y_{ti}|\bm{X}_{ti})$, where $p_{\rvw}(y|\bm{X})$ is the output of a neural network with parameters $\rvw$ that produces a distribution over target $y$ given input $\bm{X}$. We assume that $p(\rvw_t|\bm{X}_{1:t}, \rvy_{1:t}) \approx \delta(\hat{\rvw}_t - \rvw_t)$ where $\delta(\cdot)$ denotes a Dirac-delta function centered at $\rvw_t$.

The VKF algorithm at every time step is as follows:
\begin{itemize}
    \item Predict step:
    \begin{align*}
        p(\rvw_t|\bm{X}_{1:t-1}, \rvy_{1:t-1}) &= \int \delta(\hat{\rvw}_{t-1} - \rvw_{t-1})\mathcal{N}(\rvw_t|\rvw_{t-1}, \sigma^2\bm{I}) \text{d}\rvw_{t-1} \\ &= \mathcal{N}(\rvw_t|\hat{\rvw}_{t-1}, \sigma^2\bm{I})
    \end{align*}
    \item Update step:
    \begin{align*}
        p(\rvw_t|\bm{X}_{1:t}, \rvy_{1:t}) \propto \mathcal{N}(\rvw_t|\hat{\rvw}_{t-1}, \sigma^2\bm{I})\prod_{i=1}^{n_t}p_{\rvw_t}(y_{ti}|\bm{X}_{ti})
    \end{align*}
\end{itemize}
Now, we approximate $p(\rvw_t|\bm{X}_{1:t}, \rvy_{1:t})$ with a point mass $\hat{\rvw}_{t}$ such that
\begin{align*}
    \hat{\rvw}_{t} &\overset{\Delta}{=} \argmax_{\rvw_t} \left[\log{\mathcal{N}(\rvw_t|\hat{\rvw}_{t-1}, \sigma^2\bm{I})} + \sum_{i=1}^{n_t}\log{p_{\rvw_t}(y_{ti}|\bm{X}_{ti})}\right] \\
    &= \argmin_{\rvw_t} \left[\underset{\text{mean binary cross entropy}}{\underbrace{-\frac{1}{n_t}\sum_{i=1}^{n_t}\log{p_{\rvw_t}(y_{ti}|\bm{X}_{ti})}}} + \underset{\text{weight decay centered at } \hat{\rvw}_{t-1}}{\underbrace{\frac{1}{2n_t\sigma^2}\sum^D_{d=1}\left(w_{td} - \hat{w}_{t-1, d}\right)^2}}\right]
\end{align*}

In the yearbook experiment, we run 5 configurations with $\sigma^2 \in \{0.1, 0.05, 0.01, 0.005, 0.001\}$. We optimize using the same procedure as the direct fit case, with the exception that the objective has explicit regularization. 

\subsection{Variational Inference Interpretation of the Update Step}\label{sec:vi_update}

In this section, we show how truncated gradient descent on the likelihood \eqref{eq:santos_recurrence} can be interpreted as variational inference with the variational distribution \(q_t(\rvx_t) = \mathcal{N}(\rvx_t \mid \rvm_t, \rmM_t)\) where \(\rmM_t^{-1} \succ \bm{0}\) is fixed. The derivation in this section is based on Section~4.1 of \citet{khan2023bayesian}, but adapts it to the setting of the update step of a Bayesian filter and further applies the equivalence due to \citet{santos1996equivalence} in order to deduce the implied filtering covariance $\bm{\Sigma}_t^-$.  The variational lower bound on the log marginal likelihood for time step \(t\) is:
\begin{align}
\log p(\rvy_t \mid \rvy_{1:t-1}) &\ge \mathbb{E}_{q_t} \left[ \log \frac{ p(\rvx_t, \rvy_t \mid \rvy_{1:t-1}) }{ q_t(\rvx_t) } \right] \\
&= \mathbb{E}_{q_t} [\log p(\rvy_t \mid \rvx_t) + \log p(\rvx_t \mid \rvy_{1:t-1})] + \mathcal{H}(q_t)
\end{align}
Then the optimal variational distribution is:
\begin{align}
q_t^*(\rvx_t) &= \argmin_{q_t} \mathbb{E}_{q_t} [\bar{\ell}_t(\rvx_t)] - \mathcal{H}(q_t)\text{, where} \\
\bar{\ell}_t(\rvx_t) &\triangleq -\log p(\rvy_t \mid \rvx_t) - \log p(\rvx_t \mid \rvy_{1:t-1}) \\
&= -\log \mathcal{N}(\rvy_t \mid \rmH_t \rvx_t, \rmR_t) - \log \mathcal{N}(\rvx_t \mid \bm{\mu}_t^-, \bm{\Sigma}_t^-)
\end{align}
We take the approach of \citet{khan2023bayesian} where the variational optimization is performed using natural gradient descent. In particular, assume the variational distribution to be of the form
\begin{align}
q_{\bm{\lambda}_t}(\rvx_t) = h(\rvx_t) \exp[ \langle \bm{\lambda}_t, \rmT(\rvx_t) \rangle - A(\bm{\lambda}_t)]
\end{align}
Natural gradient descent on the variational objective can then be written as:
\begin{align}
\bm{\lambda}^{(k+1)}_t \leftarrow (1 - \rho^{(k)}_t) \bm{\lambda}^{(k)}_t - \rho_t^{(k)} \nabla_{\bm{\mu}} \mathbb{E}_{q_t^{(k)}}[ \bar{\ell}_t(\rvx_t) + \log h(\rvx_t)]
\end{align}
With the choice \(q_t^{(k)}(\rvx_t) = \mathcal{N}(\rvx_t \mid \rvm_t^{(k)}, \rmM_t)\), we have \(\bm{\lambda}_t^{(k)} = \rmM_t^{-1} \rvm_t^{(k)}\), \(\bm{\mu}_t^{(k)} = \rvm_t^{(k)}\) and \(2 \log h(\rvx_t) = P \log |2 \pi \rmM_t^{-1}  | - \rvx_t^\top \rmM_t^{-1} \rvx_t\). After substituting, we find that the updates can be written as follows:
\begin{align}
\rvx_t^{(k+1)} \leftarrow \rvx_t^{(k)} - \rho_t^{(k)} \rmS^{-1}_t \nabla_{\rvx_t} \bar{\ell}_t(\rvx_t) \rvert_{\rvx_t = \rvm_t^{(k)}}.
\end{align}
The gradient \(\nabla_{\rvx_t} \bar{\ell}(\rvx_t)\) takes the following form:
\begin{align}
\nabla_{\rvx_t} \bar{\ell}(\rvx_t) &= \nabla_{\rvx_t} \left[ \frac{1}{2} \| \rvy_t - \rmH_t \rvx_t \|^2_{\rmR_t} + \frac{1}{2} \| \rvx_t - \bm{\mu}_t^- \|_{\bm{\Sigma_t^-}}^2 \right] \\
&= -\rmH_t^\top \rmR_t^{-1}(\rvy_t - \rmH_t \rvx_t) + (\bm{\Sigma}^-_t)^{-1} (\rvx_t - \bm{\mu}_t^-)
\end{align}
Now if we optimize this to convergence, we would recover the setting \(\rvm_t^*\) such that \(\mathcal{N}(\rvx_t \mid \rvm_t^*, \rmM_t) \approx p(\rvx_t \mid \rvy_{1:t})\). This requires knowledge of both \(\bm{\mu}_t^-\) and \(\bm{\Sigma}_t^-\).

On the other hand, consider truncated gradient descent on the negative log-likelihood \(\ell_t(\rvx_t)\):
\begin{align}
\rvx_t^{(k+1)} &\leftarrow \rvx_t^{(k)} - \rho_t^{(k)} \rmM_t \nabla_{\rvx_t} \ell_t(\rvx_t) \rvert_{\rvx_t = \rvm_t^{(k)}}\text{, where} \\
\ell(\rvx_t) &\triangleq -\log \mathcal{N}(\rvy_t \mid \rmH_t \rvx_t, \rmR_t) \\
\nabla_{\rvx_t} \ell(\rvx_t) &= \nabla_{\rvx_t} \left[ \frac{1}{2} \| \rvy_t - \rmH_t \rvx_t \|_{\rmR_t}^2 \right] \\
&= -\rmH_t^\top \rmR_t^{-1}(\rvy_t - \rmH_t \rvx_t) \Rightarrow \\
\rvx_t^{(k+1)} &\leftarrow \rvx_t^{(k)} + \rho_t^{(k)} \rmM_t \rmH_t^\top \rmR_t^{-1}(\rvy_t - \rmH_t \rvx_t)
\end{align}
Assume that \(K\) such steps of gradient descent are taken where \(\rvx_t^{(0)} = \bm{\mu}_t^-\) and \(\rho_t^{(k)} = \rho_t\) for \(k = 0, 1, \ldots, K-1\). Then by \citet{santos1996equivalence},
\begin{align}
\rvx_t^{(K)} &= \argmin_{\rvx_t} \left[ \frac{1}{2} \| \rvy_t - \rmH_t \rvx_t \|^2_{\rmR_t} + \frac{1}{2} \| \rvx_t - \bm{\mu}_t^- \|_{\bm{\Sigma}_t^-}^2 \right],
\end{align}
where \(\bm{\Sigma}_t^-\) is defined implicitly by the choice of \(\rho_t\), \(\rmM_t\), and \(K\), in the sense we make explicit here. Let \(\rmC_t\) simultaneously diagonalize \((1 / \rho_t) \rmM_t^{-1}\) and \(\rmH_t^\top \rmR_t^{-1} \rmH_t\) (this is possible since both are symmetric and \((1 / \rho_t) \rmM_t^{-1}\) is positive definite as assumed above):
\begin{align}
\rmC_t^\top (1 / \rho_t) \rmM^{-1}_t \rmC &= \rmI \\
\rmC_t^\top \rmH_t^\top \rmR_t^{-1} \rmH_t \rmC_t &= \bm{\Lambda}_t
\end{align}
Then the equivalent \(\bm{\Sigma}_t^-\) is defined to be:
\begin{align}
\bm{\Sigma}_t^- &= \rmC_t \text{diag}( \sigma_i ) \rmC\text{, where} \\
\sigma_i &= (1 / \lambda_i) [(1 - \lambda_i)^{-k} - 1] \text{ if  } \lambda_i \neq 0 \text{ and  } 1 \text{ otherwise}.
\end{align}

\subsection{Justification Setting Measurement Noise to Identity}\label{sec:justification}

For simplicity, consider the arbitrary specification of $\bm{R}_t = \bm{I}$. In an optimization scheme over a Gaussian likelihood, this is preferable because it reduces the objective function to the mean squared error loss, which is simple to implement and fast to compute. Such an arbitrary choice implies the \emph{compensated} predict step covariance in the Kalman filter $\bm{\Sigma}_t^{-} = \bm{\Sigma}_t^{*}\bm{H}_t^\top\left(\bm{R}_t^{*}\right)^{-1}\left(\bm{H}_t^\top\right)^{+}$, where $\bm{\Sigma}_t^{*}$ is the true predict step covariance and $\bm{R}_t^{*}$ is the true measurement noise at timestep $t$. The Moore–Penrose inverse of matrix $\bm{A}$ is denoted $\bm{A}^+$. 

\textbf{Proof:} Let $\bm{\Sigma}_t^{*}$ be the true predict step covariance and $\bm{R}_t^{*}$ be the true measurement noise at timestep $t$. Suppose we wish to identify the quantity of $\bm{X}$ that makes the following expression true:
        \begin{align*}
            \bm{\Sigma}_t^{*}\bm{H}_t^\top\left(\bm{H}_t\bm{\Sigma}_t^{*}\bm{H}_t^\top + \bm{R}_t^{*}\right)^{-1} = \bm{X}\bm{H}_t^\top\left(\bm{H}_t\bm{X}\bm{H}_t^\top + \bm{I}_M\right)^{-1}.
        \end{align*}
        This is exactly our Kalman gain expression where the L.H.S represents the optimal estimate and the R.H.S shows an arbitrary matrix $\bm{X}$ given $\bm{R}_t = \bm{I}_M$.
        \begin{align*}
            \bm{\Sigma}_t^{*}\bm{H}_t^\top\left(\bm{H}_t\bm{\Sigma}_t^{*}\bm{H}_t^\top + \bm{R}_t^{*}\right)^{-1} = \bm{X}\bm{H}_t^\top\left(\bm{H}_t\bm{X}\bm{H}_t^\top + \bm{I}_M\right)^{-1} \\
             \bm{\Sigma}_t^{*}\bm{H}_t^\top\left(\bm{H}_t\bm{\Sigma}_t^{*}\bm{H}_t^\top + \bm{R}_t^{*}\right)^{-1}\bm{H}_t\bm{X}\bm{H}_t^\top + \bm{\Sigma}_t^{*}\bm{H}_t^\top\left(\bm{H}_t\bm{\Sigma}_t^{*}\bm{H}_t^\top + \bm{R}_t^{*}\right)^{-1} = \bm{X}\bm{H}_t^\top \\
             \left(\bm{H}_t\bm{\Sigma}_t^{*}\bm{H}_t^\top + \bm{R}_t^{*}\right)^{-1}\bm{H}_t\bm{X}\bm{H}_t^\top + \left(\bm{H}_t\bm{\Sigma}_t^{*}\bm{H}_t^\top + \bm{R}_t^{*}\right)^{-1} = \left(\bm{H}_t^\top\right)^{+}\left(\bm{\Sigma}_t^{*}\right)^{-1}\bm{X}\bm{H}_t^\top \\
             \bm{H}_t\bm{X}\bm{H}_t^\top + \bm{I}_M = \bm{H}_t\bm{\Sigma}_t^{*}\bm{H}_t^\top\left(\bm{H}_t^\top\right)^{+}\left(\bm{\Sigma}_t^{*}\right)^{-1}\bm{X}\bm{H}_t^\top + \bm{R}_t^{*}\left(\bm{H}_t^\top\right)^{+}\left(\bm{\Sigma}_t^{*}\right)^{-1}\bm{X}\bm{H}_t^\top \\
             \bm{H}_t\bm{X}\bm{H}_t^\top + \bm{I}_M = \bm{H}_t\bm{X}\bm{H}_t^\top + \bm{R}_t^{*}\left(\bm{H}_t^\top\right)^{+}\left(\bm{\Sigma}_t^{*}\right)^{-1}\bm{X}\bm{H}_t^\top \\
             \bm{X} = \bm{\Sigma}_t^{*}\bm{H}_t^\top\left(\bm{R}_t^{*}\right)^{-1}\left(\bm{H}_t^\top\right)^{+} = \bm{\Sigma}_t^{-} \\
             \square
        \end{align*}

Such a specification considerably simplifies computation in an \emph{implicit} scheme. By setting $\bm{R}_t = \bm{I}$, we do not need to store $\bm{R}_t$ or perform any computations with it. We just need to pick an optimizer that correctly specifies this \emph{compensated} predict step covariance. The argument extends to the nonlinear case with only minor adjustments.

\subsection{Grid Search Details}\label{sec:grid}

The framework proposed in this paper attempts to define Bayesian filtering equations \emph{implicitly} via specifying an appropriate optimizer. This requires a small validation set and a hyperparameter search, which we will now describe in the context of our experiments. We also describe the grid search performed for the extended Kalman filter (EKF) and unscented Kalman filter (UKF) in Section~\ref{sec:lorenz}.

\subsubsection{Toy nonlinear System \& Stochastic Lorenz Attractor}\label{sec:grid1}

In both the toy nonlinear system and the stochastic Lorenz attactor system, we perform a grid search for Adam \citep{kingma2014adam}, RMSprop \citep{tieleman2012lecture}, Adagrad \citep{duchi2011adaptive}, Adadelta \citep{zeiler2012adadelta}, and gradient descent. For 5 separate Monte Carlo (MC) experiments, we test every combination of step size $K \in \{1, 3, 5, 10, 25, 50, 100\}$, learning rate $\eta \in \{1.0, 0.5, 0.1, 0.05, 0.01\}$, and decay terms $\gamma, \beta_1, \beta_2 \in \{0.1, 0.5, 0.9\}$ where applicable. Adam's decay terms are $\beta_1$ and $\beta_2$ whereas $\gamma$ only applies to RMSprop. For Adam, we always set $\beta_2 = \beta_1$ to simplify the search. In total, this corresponds to 7 test configurations for Adadelta, 35 test configurations each for gradient descent and Adagrad, and 105 test configurations each for RMSprop and Adam.

For only the stochastic Lorenz Attractor, we also perform a grid search for the process noise covariance $\bm{Q}_{t-1}$. The process noise covariance $\bm{Q}_{t-1}$ is determined by a scalar $\alpha$ that defines the spectral density of a three-dimensional Wiener process. We test four systems in this paper where $\alpha = 1, 5, 10, 20$. To optimize $\bm{Q}_{t-1}$, which is determined by $\alpha$, we perform a grid search over 500 evenly spaced values of $\alpha \in [0.5, 250]$ for both the EKF and UKF in all four experiments. Comparatively, we test roughly 14 times the number of EKF configurations than the Implicit MAP Filter with gradient descent and roughly 2 times more configurations than all Implicit MAP Filters combined.

\subsubsection{Yearbook}\label{sec:grid2}

For this experiment, we reduce the grid search to only five configurations of Adam where all hyperparameters are set to standard settings ($\eta = 0.001, \beta_1 = 0.9, \beta_2 = 0.999$) and the number of steps $K$ is selected from the set $\{1, 10, 25, 50, 100\}$. As described in Appendix \ref{sec:extended3}, we divide up the 80 filtering years of the yearbook dataset into 40 tuning years and 40 testing years. During the tuning years, we have a small held-out validation set of 16 examples for each year that we use to calculate the classification accuracy for the parameters found by each optimizer. In the main paper, we report the optimizer that performs the best on this validation set.

We similarly perform a grid search for the variational Kalman filter (VKF) and particle filter (PF) for the $\sigma^2$ term in the transition distribution. Since the transition distribution is unknown, a grid search is required. We report the results for the VKF and PF that perform the best on the same validation set, with $\sigma^2 \in \{0.1, 0.05, 0.01, 0.005, 0.001\}$.

\subsection{Extended Experimental Results and Details}\label{sec:extended}

We report additional results, tables, and figures from all three experiments in this section. For the yearbook dataset, we give a complete description of the experiment performed.

\subsubsection{Toy nonlinear System}\label{sec:extended1}

In the main paper, we report results for 100 Monte Carlo (MC) experiments from the \emph{best} optimizer hyperparameters found using a grid search over 5 separate MC experiments. To give the reader some intuition about good hyperparameters in this system, we show the top 20 performing optimizers from the grid search in Table \ref{supp_1_table} ($\bm{Q}_{t-1} = 3, \bm{R}_t = 2$). 287 configurations are tested in total across Adam \citep{kingma2014adam}, RMSprop \citep{tieleman2012lecture}, Adadelta \citep{zeiler2012adadelta}, Adagrad \citep{duchi2011adaptive}, and gradient descent. Of those 287 configurations, the top 20 performing configurations are exclusively between Adam and RMSprop.

\begin{table}[ht]
\caption{Top 20 optimizers found in grid search for the toy nonlinear system ($\bm{Q}_{t-1} = 3, \bm{R}_t = 2$). The RMSprop and Adam configurations reported in the main paper are shown in \textbf{bold}. $K$ denotes the number of steps used at every time step. $\eta$ is the learning rate. $\gamma, \beta_1, \beta_2$ are the decay terms specific to each optimizer.}
\label{supp_1_table}
\centering
\renewcommand{\arraystretch}{1.0} %
\begin{tabular}{lc}
\toprule
\textbf{Method} & \textbf{RMSE} \\
\midrule
RMSprop ($K = 5, \eta = 1.0, \gamma = 0.5$) & $6.391 \pm 0.230$  \\
RMSprop ($K = 50, \eta = 0.1, \gamma = 0.9$) & $6.380 \pm 0.230$  \\
RMSprop ($K = 10, \eta = 0.5, \gamma = 0.1$) & $6.293 \pm 0.222$  \\
RMSprop ($K = 10, \eta = 0.5, \gamma = 0.5$) & $6.248 \pm 0.240$  \\
Adam ($K = 10, \eta = 0.5, \beta_1, \beta_2 = 0.9$) & $6.205 \pm 0.241$ \\
RMSprop ($K = 100, \eta = 0.05, \gamma = 0.9$) & $6.201 \pm 0.236$  \\
Adam ($K = 5, \eta = 1.0, \beta_1, \beta_2 = 0.9$) & $6.112 \pm 0.239$ \\
Adam ($K = 10, \eta = 0.5, \beta_1, \beta_2 = 0.1$) & $6.059 \pm 0.221$ \\
Adam ($K = 25, \eta = 0.1, \beta_1, \beta_2 = 0.9$) & $6.027 \pm 0.238$ \\
RMSprop ($K = 50, \eta = 0.1, \gamma = 0.5$) & $6.018 \pm 0.223$  \\
RMSprop ($K = 100, \eta = 0.05, \gamma = 0.5$) & $6.010 \pm 0.229$  \\
\textbf{RMSprop} (\bm{$K = 50, \eta = 0.1, \gamma = 0.1$}) & $\bm{6.000 \pm 0.227}$  \\
RMSprop ($K = 100, \eta = 0.05, \gamma = 0.1$) & $5.987 \pm 0.225$  \\
Adam ($K = 5, \eta = 1.0, \beta_1, \beta_2 = 0.5$) & $5.973 \pm 0.218$ \\
Adam ($K = 50, \eta = 0.05, \beta_1, \beta_2 = 0.9$) & $5.963 \pm 0.224$ \\
Adam ($K = 10, \eta = 0.5, \beta_1, \beta_2 = 0.5$) & $5.953 \pm 0.219$ \\
\textbf{Adam} (\bm{$K = 50, \eta = 0.1, \beta_1, \beta_2 = 0.1$}) & $\bm{5.842 \pm 0.231}$ \\
Adam ($K = 100, \eta = 0.05, \beta_1, \beta_2 = 0.1$) & $5.830 \pm 0.232$ \\
Adam ($K = 50, \eta = 0.1, \beta_1, \beta_2 = 0.5$) & $5.794 \pm 0.216$ \\
Adam ($K = 100, \eta = 0.05, \beta_1, \beta_2 = 0.5$) & $5.780 \pm 0.220$ \\
\bottomrule
\end{tabular}
\end{table}

From Table \ref{supp_1_table}, it is clear that there is a diverse set of optimizer hyperparameters that can produce good results. The goal of hyperparameter selection in this context is to strike a balance between overfitting and underfitting the likelihood such that it reflects a similar balance between the prior and measurement noise in the explicit filtering sense. It is clear that momentum plays a beneficial role here, since gradient descent and the EKF do not perform as well as optimizers with momentum. 

In the main paper, we reported results where the true process noise $\bm{Q}_{t-1} = 1, 3, 5$ and the true measurement noise $\bm{R}_{t} = 2$. In Table \ref{experiment_1_table1}, we show results for $\bm{Q}_{t-1} = 1, 3, 5$ and $\bm{R}_{t} = 1$. In Table \ref{experiment_1_table3}, we show results for $\bm{Q}_{t-1} = 1, 3, 5$ and $\bm{R}_{t} = 3$. This demonstrates 6 additional configurations of the toy nonlinear system, each showing a similar result. In low to medium process noise settings, the Implicit MAP Filter is comparable to the Unscented Kalman Filter (UKF). In high process noise settings, the UKF is outperformed by the Implicit MAP Filter. The particle filter (PF) performs the best universally. The extended Kalman filter (EKF) and iterated extended Kalman filter (IEKF) diverged under all settings. 

In summary, Table \ref{experiment_1_table}, Table \ref{experiment_1_table1}, and Table \ref{experiment_1_table3} show that the PF outperforms our Implicit MAP Filter in 9 out of 9 experiments, the UKF outperforms the Implicit MAP Filter in 5 out 9 experiments (with performance being relatively comparable across all 9 experiments), and the EKF/IEKF outperforms the Implicit MAP Filter in 0 out of 9 experiments. 

\begin{table}[t]
\caption{RMSEs on the toy nonlinear system ($\bm{R}_t = 1$). Results show the average RMSE over 100 MC simulations with 95\% confidence intervals.}
\label{experiment_1_table1}
\centering
\renewcommand{\arraystretch}{1.0} %
\begin{tabular}{lccc}
\toprule
\textbf{Method} & \textbf{RMSE ($\bm{Q}_{t-1} = 1$)} & \textbf{RMSE ($\bm{Q}_{t-1} = 3$)} & \textbf{RMSE ($\bm{Q}_{t-1} = 5$)} \\
\midrule
EKF & $29.188 \pm 5.335$ & $38.075 \pm 5.564$ & $45.102 \pm 3.876$ \\
IEKF ($K = 5$) & $11.252 \pm 0.643$ & $17.071 \pm 0.453$ & $19.203 \pm 0.494$ \\
\midrule
IMAP (Adadelta)* & $30.984 \pm 5.688$ & $33.600 \pm 8.007$ & $21.210 \pm 3.865$ \\
IMAP (Gradient Descent)* & $5.564 \pm 0.224$ & $7.931 \pm 0.159$ & $10.142 \pm 0.172$ \\
IMAP (Adagrad)* & $5.181 \pm 0.225$ & $6.549 \pm 0.223$ & $9.446 \pm 0.230$ \\
IMAP (RMSprop)* & $5.138 \pm 0.219$ & $5.966 \pm 0.224$ & $8.829 \pm 0.264$ \\
IMAP (Adam)* & $5.109 \pm 0.201$ & $5.708 \pm 0.239$ & $8.341 \pm 0.315$ \\
\midrule
UKF & $4.178 \pm 0.274$ & $6.572 \pm 0.392$ & $12.735 \pm 0.616$ \\
\midrule
PF ($n = 1000$) & $1.263 \pm 0.043$ & $2.406 \pm 0.128$ & $4.264 \pm 0.161$ \\
\bottomrule
\end{tabular}
*Methods where the reported hyperparameters were found via grid search (see Appendix \ref{sec:grid}).
\end{table}

\begin{table}[t]
\caption{RMSEs on the toy nonlinear system ($\bm{R}_t = 3$). Results show the average RMSE over 100 MC simulations with 95\% confidence intervals.}
\label{experiment_1_table3}
\centering
\renewcommand{\arraystretch}{1.0} %
\begin{tabular}{lccc}
\toprule
\textbf{Method} & \textbf{RMSE ($\bm{Q}_{t-1} = 1$)} & \textbf{RMSE ($\bm{Q}_{t-1} = 3$)} & \textbf{RMSE ($\bm{Q}_{t-1} = 5$)} \\
\midrule
EKF & $24.019 \pm 3.322$ & $34.099 \pm 3.706$ & $39.908 \pm 4.277$ \\
IEKF ($K = 5$) & $8.756 \pm 0.606$ & $13.860 \pm 0.494$ & $17.252 \pm 0.510$ \\
\midrule
IMAP (Adadelta)* & $37.356 \pm 12.733$ & $26.783 \pm 6.720$ & $28.075 \pm 11.094$ \\
IMAP (Gradient Descent)* & $5.714 \pm 0.251$ & $8.099 \pm 0.163$ & $10.082 \pm 0.184$ \\
IMAP (Adagrad)* & $5.781 \pm 0.224$ & $6.630 \pm 0.213$ & $9.349 \pm 0.238$ \\
IMAP (RMSprop)* & $5.444 \pm 0.254$ & $6.120 \pm 0.224$ & $8.300 \pm 0.408$ \\
IMAP (Adam)* & $6.045 \pm 0.176$ & $5.978 \pm 0.209$ & $7.865 \pm 0.307$ \\
\midrule
UKF & $4.720 \pm 0.250$ & $5.665 \pm 0.230$ & $8.490 \pm 0.334$ \\
\midrule
PF ($n = 1000$) & $1.815 \pm 0.055$ & $3.153 \pm 0.104$ & $4.757 \pm 0.152$ \\
\bottomrule
\end{tabular}
*Methods where the reported hyperparameters were found via grid search (see Appendix \ref{sec:grid}).
\end{table}

\begin{figure}[h]
\label{fig:histo}
\begin{center}
\includegraphics[width=\textwidth]{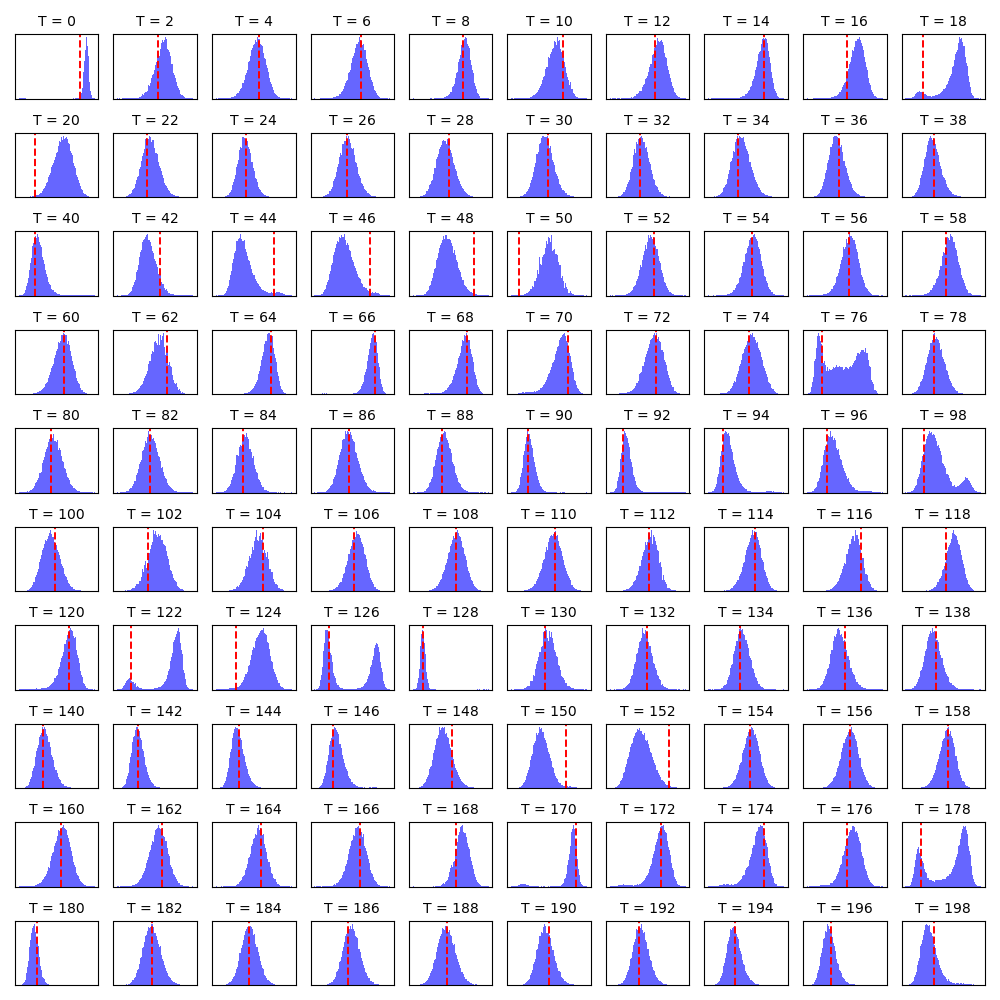}
\end{center}
\caption{Filtering distribution from the particle filter (blue) for $100000$ particles and the MAP estimates from the Implicit MAP Filter (red) using a single random experiment ($\bm{Q}_{t-1} = 3, \bm{R}_t = 2$).}
\end{figure}

In Figure 3, we take a random experimental seed (for $\bm{Q}_{t-1} = 3, \bm{R}_t = 2$) and show the filtering distributions from every other time step produced by a particle filter with $100,000$ particles. On top of the filtering distributions, we show the maximum a posteriori (MAP) estimates produced by Adam with gradient steps $n = 50$, learning rate $\eta = 0.1$, and momentum terms $\beta_1, \beta_2 = 0.1$ (red). On most time steps, our Implicit MAP Filter exactly maximizes the filtering distribution maintained by the particle filter, despite all time steps being non-convex optimization problems. The trials where the Implicit MAP Filter estimates do not correspond to the true MAP most often come on transition trials, where the state is making an aggressive crossing over the barrier in the double-well. These are the trials where the EKF tends to diverge and the UKF equivalently struggles to produce accurate estimates.

In Figure 4, we visualize the update steps of both an EKF and an Implicit MAP Filter with an Adam optimizer ($K = 25, \eta = 0.1, \beta_1 = 0.9, \beta_2 = 0.9$). Both of these methods rely on similar gradient information so it is important to understand why the EKF fails catastrophically and diverges but adaptive optimizers, such as Adam, do not suffer the same effect in this system. This figure sheds some light on the situation, showing that the toy nonlinear system can be broken into three different modes: pre-transition, transition, and post-transition. At time step $t = 5$, we are in the pre-transition stage where the objective is approximately convex within the local optimization region. Both Adam and the EKF produce similar estimates in this pre-transition stage. At time steps $t = 12$ and $t = 15$, we are entering the transition phase where the double well shape of the likelihood begins to flatten out. Time step $t = 16$ is the most important step. Here, the update step must cross the barrier at 0 in order to avoid divergence. Adam crosses the barrier at 0 because its incorporation of momentum, which continues to move the estimates despite there being a lack of gradient. The EKF update step does not have momentum. At time step $t = 17$, post-transition, the EKF finds itself on the wrong side of the double well. Since measurements occur only on one side of the double well, as shown at time step $T = 27$, the EKF cannot recover, hence the divergence. 

\begin{figure}[h]
\label{gradients}
\begin{center}
\includegraphics[width=\textwidth]{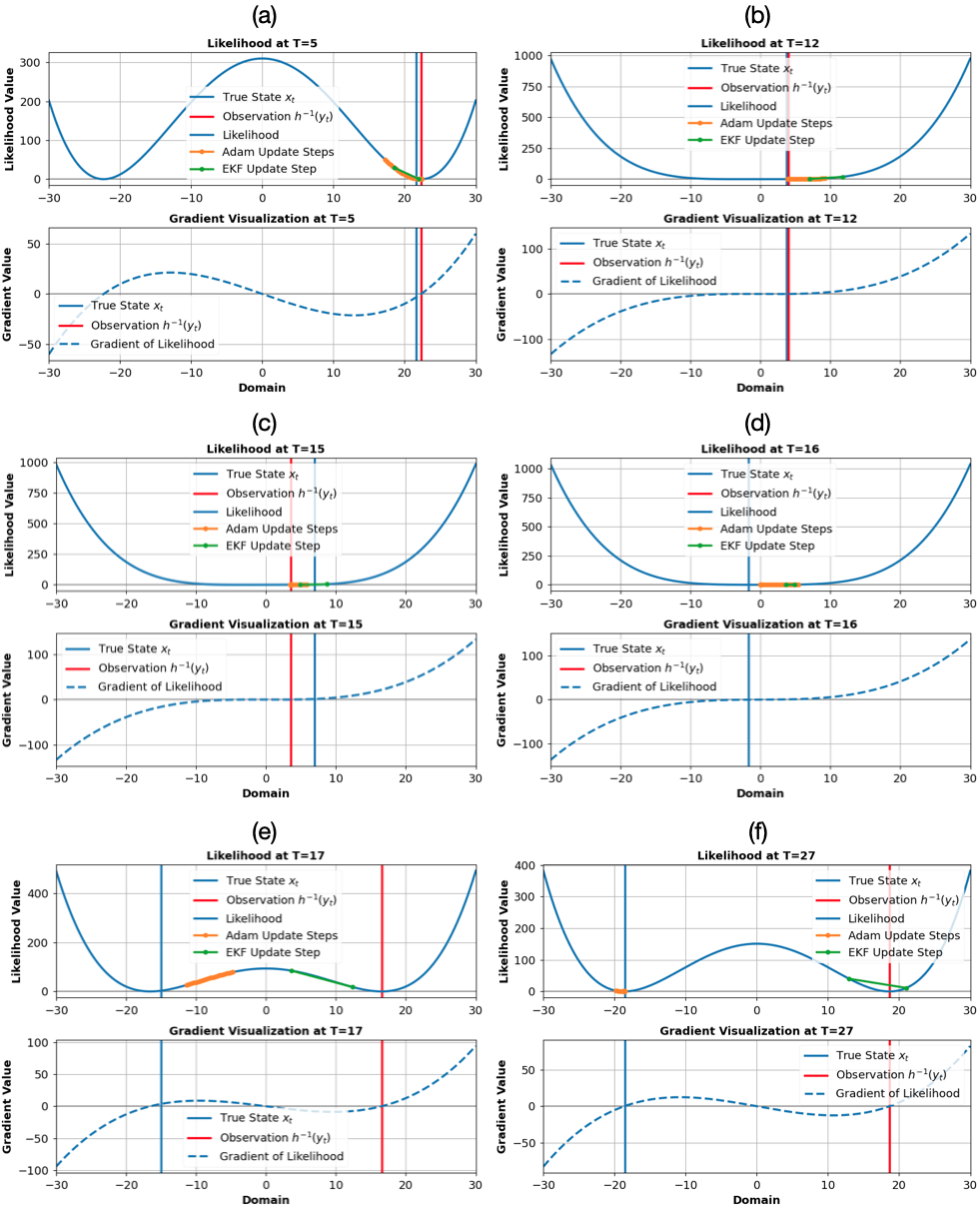}
\end{center}
\caption{Likelihood and gradient visualization for the toy nonlinear system. (a) pre-transition time steps, (b,c,d) transition time steps. (e, f) post-transition time steps. In (a), the objective function is locally convex and gradient information can be used effectively. The likelihood takes the shape of a double-well. In (b, c, d), gradients approach zero as the double-well flattens out. In (e,f), the double-well picks up again, requiring estimates to be on the correct side of the double-well in order to perform close to optimal. Experiment was run with a random seed and $\bm{Q}_{t-1} = 3, \bm{R}_t = 2$. Other random seeds produce nearly identical observations.}
\end{figure}

\begin{table}[h]
\caption{The effect of modifying the number of gradient steps using the optimizer configuration reported in the main section for Adam on the toy nonlinear system. Ordered by RMSE.}
\label{supp_3_table}
\centering
\renewcommand{\arraystretch}{1.0} %
\begin{tabular}{lc}
\toprule
\textbf{Method} & \textbf{RMSE} \\
\midrule
Adam ($K = 1, \eta = 0.1, \beta_1, \beta_2 = 0.1$) & $10.510 \pm 0.263$  \\
Adam ($K = 25, \eta = 0.1, \beta_1, \beta_2 = 0.1$) & $10.478 \pm 0.409$  \\
Adam ($K = 3, \eta = 0.1, \beta_1, \beta_2 = 0.1$) & $9.749 \pm 0.288$  \\
Adam ($K = 5, \eta = 0.1, \beta_1, \beta_2 = 0.1$) & $9.244 \pm 0.266$  \\
Adam ($K = 10, \eta = 0.1, \beta_1, \beta_2 = 0.1$) & $9.218 \pm 0.291$  \\
Adam ($K = 100, \eta = 0.1, \beta_1, \beta_2 = 0.1$) & $6.575 \pm 0.347$  \\
Adam ($K = 50, \eta = 0.1, \beta_1, \beta_2 = 0.1$) & $5.842 \pm 0.231$  \\
\bottomrule
\end{tabular}
\end{table}

Finally, in Table \ref{supp_3_table}, we show the effect of changing the number of gradient steps for the Adam optimizer we report in the main section, holding all other hyperparameters fixed. The number of gradient steps, $K$, is the most natural hyperparameter to adjust, and Table \ref{supp_3_table} demonstrates that the specification of $K$ plays a strong role in performance when attempting to implicitly filter in a time-varying setting.

\subsubsection{Stochastic Lorenz Attractor}\label{sec:extended2}

Similar to the previous section, we report the top 20 optimizers found from the grid search in Table \ref{supp_2_table} using the ideal condition (RK4) ($\alpha = 10$). Exactly as before, we tested 287 configurations in total across Adam, RMSprop, Adadelta, Adagrad, and gradient descent. Of those 287 configurations, we achieve a more diverse set of optimizers that performed well compared to the previous system.

\begin{table}[ht]
\caption{Top 20 optimizers found in grid search for the stochastic Lorenz attractor ($\alpha = 10$) with respect to the RK4 case. We also show performance of the same optimizer configurations for Euler and GRW cases. RMSEs that outperform the EKF (original $\bm{Q}_{t-1}$) are shown in \textbf{bold}. The same configuration is robust from RK4 $\to$ Euler. GRW requires different hyperparameters for the optimizer since the top RK4 optimizers do not correspond to top GRW optimizers.}
\label{supp_2_table}
\centering
\renewcommand{\arraystretch}{1.0} %
\begin{tabular}{lccc}
\toprule
\textbf{Method} & \textbf{RMSE (RK4)} & \textbf{RMSE (Euler)} & \textbf{RMSE (GRW)} \\
\midrule
Adam ($K = 10, \eta = 0.05, \beta_1, \beta_2 = 0.5$) & $0.903 \pm 0.014$ & $\bm{1.189 \pm 0.025}$ & $6.728 \pm 0.101$  \\
RMSprop ($K = 10, \eta = 0.05, \gamma = 0.5$) & $0.903 \pm 0.112$ & $\bm{1.525 \pm 0.009}$ & $7.526 \pm 0.105$  \\
RMSprop ($K = 50, \eta = 0.01, \gamma = 0.1$) & $0.900 \pm 0.079$ & $\bm{1.470 \pm 0.111}$ & $7.516 \pm 0.105$  \\
RMSprop ($K = 50, \eta = 0.01, \gamma = 0.5$) & $0.896 \pm 0.077$ & $\bm{1.473 \pm 0.111}$ & $7.518 \pm 0.105$  \\
Adam ($K = 25, \eta = 0.01, \beta_1, \beta_2 = 0.9$) & $0.895 \pm 0.015$ & $\bm{1.221 \pm 0.026}$ & $8.335 \pm 0.108$ \\
Adam ($K = 5, \eta = 0.1, \beta_1, \beta_2 = 0.5$) & $0.893 \pm 0.014$ & $\bm{1.219 \pm 0.026}$ & $6.896 \pm 0.102$  \\
RMSprop ($K = 5, \eta = 0.1, \gamma = 0.5$) & $0.893 \pm 0.102$ & $\bm{1.118 \pm 0.015}$ & $7.531 \pm 0.105$ \\
RMSprop ($K = 10, \eta = 0.05, \gamma = 0.1$) & $0.890 \pm 0.064$ & $\bm{1.472 \pm 0.112}$ & $7.517 \pm 0.105$ \\
Gradient Descent ($K = 10, \eta = 0.01$) & $\bm{0.888 \pm 0.090}$ & $\bm{2.286 \pm 0.148}$ & $5.962 \pm 0.088$ \\
Adam ($K = 5, \eta = 0.1, \beta_1, \beta_2 = 0.1$) & $\bm{0.882 \pm 0.049}$ & $\bm{1.416 \pm 0.105}$ & $7.411 \pm 0.104$  \\
RMSprop ($K = 10, \eta = 0.05, \gamma = 0.9$) & $\bm{0.881 \pm 0.114}$ & $\bm{1.540 \pm 0.123}$ & $7.554 \pm 0.104$  \\
RMSprop ($K = 5, \eta = 0.1, \gamma = 0.9$) & $\bm{0.881 \pm 0.113}$ & $\bm{1.545 \pm 0.124}$ & $7.555 \pm 0.104$  \\
Adam ($K = 10, \eta = 0.05, \beta_1, \beta_2 = 0.1$) & $\bm{0.877 \pm 0.039}$ & $\bm{1.417 \pm 0.122}$ & $7.395 \pm 0.104$  \\
Adam ($K = 50, \eta = 0.01, \beta_1, \beta_2 = 0.1$) & $\bm{0.876 \pm 0.037}$ & $\bm{1.391 \pm 0.083}$ & $7.384 \pm 0.104$ \\
Gradient Descent ($K = 1, \eta = 0.1$) & $\bm{0.849 \pm 0.077}$ & $\bm{2.114 \pm 0.125}$ & $5.852 \pm 0.086$ \\
Adagrad ($K = 50, \eta = 0.05$) & $\bm{0.846 \pm 0.014}$ & $\bm{1.213 \pm 0.027}$ & $6.905 \pm 0.102$ \\
Gradient Descent ($K = 3, \eta = 0.1$) & $\bm{0.799 \pm 0.009}$ & $\bm{0.960 \pm 0.012}$ & $\bm{3.061 \pm 0.038}$ \\
Gradient Descent ($K = 5, \eta = 0.05$) & $\bm{0.743 \pm 0.010}$ & $\bm{0.996 \pm 0.014}$ & $3.575 \pm 0.046$ \\
Gradient Descent ($K = 25, \eta = 0.01$) & $\bm{0.738 \pm 0.010}$ & $\bm{1.002 \pm 0.015}$ & $3.627 \pm 0.047$ \\
Gradient Descent ($K = 3, \eta = 0.05$) & $\bm{0.701 \pm 0.018}$ & $\bm{1.316 \pm 0.027}$ & $4.911 \pm 0.068$ \\
\bottomrule
\end{tabular}
\end{table}

It is clear that the same optimizers perform well when the accuracy of the numerical integration degrades. The number of optimizers that outperform the EKF (original $\bm{Q}_{t-1}$) increases as we move from 4th order Runge-Kutta (RK4) to Euler's Method. This reflects a robustness property that is desirable in this implicit formulation. When numerical integration is completely removed, as it is for the Gaussian random walk, most of the optimizers that work well with RK4 do not work well here, demonstrating that the Implicit MAP Filter is not perfectly robust to dynamics misspecification. This result is sensible, as the transition distribution itself changed, not just the numerical approximation of it.

\begin{figure}[h]
\label{lorenz_8plots}
\begin{center}
\includegraphics[width=\textwidth]{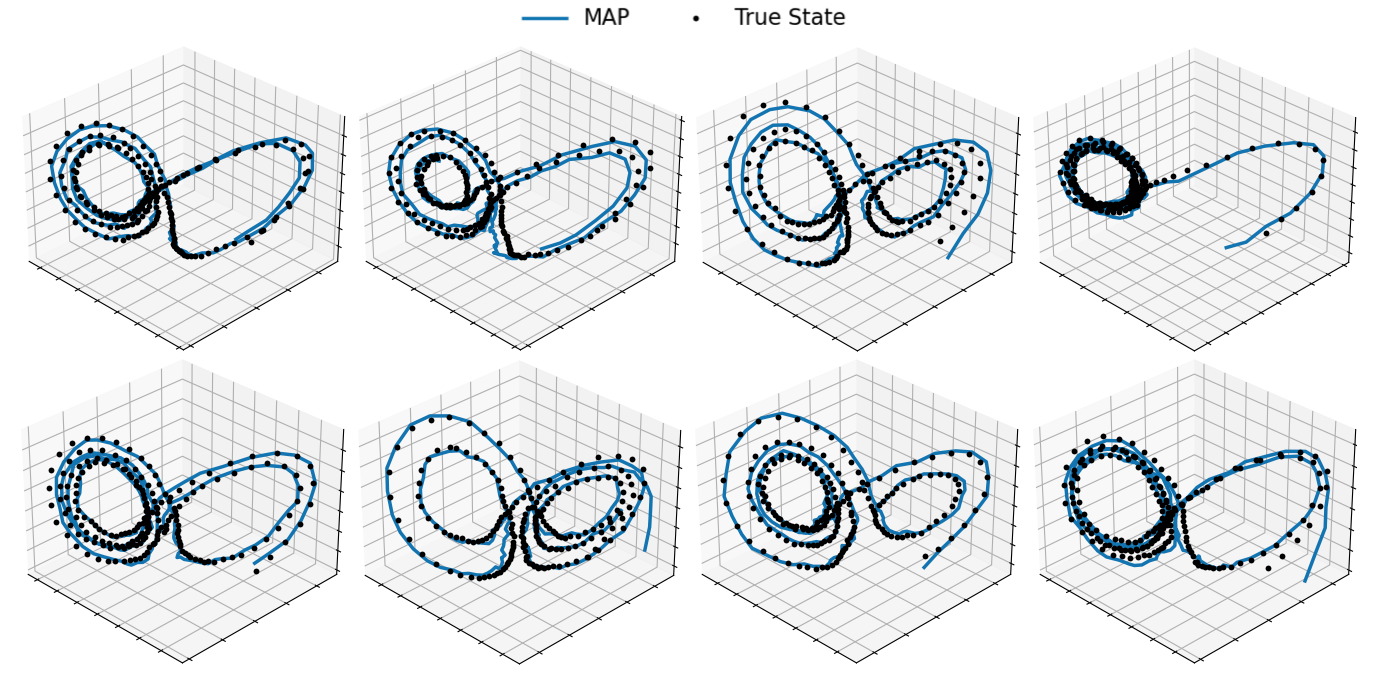}
\end{center}
\caption{Eight random trajectories for the stochastic Lorenz attractor with 4th order Runge-Kutta and the filter estimates for IMAP with gradient descent. 3 steps with learning rate $\eta = 0.05$.}
\end{figure}

In Figure 5, we show the fitted trajectories for 8 different stochastic Lorenz attractor experiments using gradient descent with 3 steps and learning rate $\eta = 0.05$. Two things are important to note: (i) the stochastic Lorenz attractor has extreme variation in the trajectories it produces and (ii) gradient descent does a remarkably good job of fitting this system despite the stochastic differential equation being chaotic. The extreme variation in the trajectories makes every random seed a different filtering problem.

\begin{table}[ht]
\caption{The effect of modifying the number of gradient steps using the optimizer configuration reported in the main section for Gradient Descent on the stochastic Lorenz attractor (RK4). Ordered by RMSE.}
\label{supp_4_table}
\centering
\renewcommand{\arraystretch}{1.0} %
\begin{tabular}{lc}
\toprule
\textbf{Method} & \textbf{RMSE} \\
\midrule
Gradient Descent ($K = 100, \eta = 0.05$) & $1.985 \pm 0.011$  \\
Gradient Descent ($K = 1, \eta = 0.05$) & $1.937 \pm 0.282$  \\
Gradient Descent ($K = 50, \eta = 0.05$) & $1.847 \pm 0.010$  \\
Gradient Descent ($K = 25, \eta = 0.05$) & $1.493 \pm 0.009$  \\
Gradient Descent ($K = 10, \eta = 0.05$) & $0.987 \pm 0.008$  \\
Gradient Descent ($K = 5, \eta = 0.05$) & $0.743 \pm 0.010$  \\
Gradient Descent ($K = 3, \eta = 0.05$) & $0.701 \pm 0.018$  \\
\bottomrule
\end{tabular}
\end{table}

\begin{table}[ht]
\caption{RMSEs on the stochastic Lorenz attractor ($\alpha = 1$). Results show the average RMSE over 100 MC simulations with 95\% confidence intervals. RK4 indicates 4th order Runge-Kutta method. Euler indicates Euler's method. GRW indicates a Gaussian random walk.}
\label{experiment_2_q1}
\centering
\renewcommand{\arraystretch}{1.0} %
\begin{tabular}{lccc}
\toprule
\textbf{Method} & \textbf{RMSE (RK4)} & \textbf{RMSE (Euler)} & \textbf{RMSE (GRW)} \\
\midrule
EKF (original $\bm{Q}_{t-1}$) & $4.948 \pm 0.600$ & $8.940 \pm 0.310$ & $5.953 \pm 0.041$ \\
EKF (optimized $\bm{Q}_{t-1}$)* & $\bm{0.621 \pm 0.028}$ & $\bm{0.939 \pm 0.011}$ & $\bm{1.568 \pm 0.011}$ \\
IEKF ($K = 5$) & $4.948 \pm 0.600$ & $8.940 \pm 0.310$ & $5.953 \pm 0.041$ \\
\midrule
IMAP (Adadelta)* & $6.480 \pm 0.450$ & $7.713 \pm 0.441$ & $10.774 \pm 0.216$ \\
IMAP (Adam)* & $0.845 \pm 0.035$ & $1.168 \pm 0.014$ & $1.897 \pm 0.011$ \\
IMAP (Adagrad)* & $0.824 \pm 0.067$ & $1.098 \pm 0.017$ & $1.797 \pm 0.010$  \\
IMAP (RMSprop)* & $0.823 \pm 0.082$ & $1.076 \pm 0.014$ & $1.752 \pm 0.015$ \\
IMAP (Gradient Descent)* & $\bm{0.626 \pm 0.031}$ & $\bm{0.947 \pm 0.011}$ & $\bm{1.568 \pm 0.011}$  \\
\midrule
UKF (original $\bm{Q}_{t-1}$) & $5.834 \pm 0.251$ & $6.232 \pm 0.255$ & $9.398 \pm 0.183$  \\
UKF (optimized $\bm{Q}_{t-1}$)* & $1.396 \pm 0.011$ & $1.410 \pm 0.011$ & $1.738 \pm 0.012$  \\
\midrule
PF ($n = 1000$) & $1.461 \pm 0.021$ & $1.725 \pm 0.031$ & $15.280 \pm 0.232$  \\
\bottomrule
\end{tabular}

*Methods where the reported hyperparameters were found via grid search (see Appendix \ref{sec:grid}).
\end{table}

\begin{table}[ht]
\caption{RMSEs on the stochastic Lorenz attractor ($\alpha = 5$). Results show the average RMSE over 100 MC simulations with 95\% confidence intervals. RK4 indicates 4th order Runge-Kutta method. Euler indicates Euler's method. GRW indicates a Gaussian random walk.}
\label{experiment_2_q5}
\centering
\renewcommand{\arraystretch}{1.0} %
\begin{tabular}{lccc}
\toprule
\textbf{Method} & \textbf{RMSE (RK4)} & \textbf{RMSE (Euler)} & \textbf{RMSE (GRW)} \\
\midrule
EKF (original $\bm{Q}_{t-1}$) & $1.510 \pm 0.305$ & $5.430 \pm 0.302$ & $3.954 \pm 0.032$ \\
EKF (optimized $\bm{Q}_{t-1}$)* & $\bm{0.651 \pm 0.025}$ & $\bm{0.938 \pm 0.012}$ & $\bm{1.564 \pm 0.010}$ \\
IEKF ($K = 5$) & $1.510 \pm 0.305$ & $5.430 \pm 0.302$ & $3.954 \pm 0.032$ \\
\midrule
IMAP (Adadelta)* & $6.212 \pm 0.424$ & $7.560 \pm 0.477$ & $10.736 \pm 0.232$ \\
IMAP (Adam)* & $0.870 \pm 0.015$ & $1.163 \pm 0.021$ & $1.899 \pm 0.011$ \\
IMAP (Adagrad)* & $0.843 \pm 0.114$ & $1.126 \pm 0.011$ & $1.797 \pm 0.010$  \\
IMAP (RMSprop)* & $0.835 \pm 0.119$ & $1.076 \pm 0.016$ & $1.755 \pm 0.0165$ \\
IMAP (Gradient Descent)* & $\bm{0.665 \pm 0.045}$ & $\bm{0.947 \pm 0.012}$ & $\bm{1.566 \pm 0.010}$  \\
\midrule
UKF (original $\bm{Q}_{t-1}$) & $4.066 \pm 0.124$ & $4.208 \pm 0.127$ & $7.106 \pm 0.083$  \\
UKF (optimized $\bm{Q}_{t-1}$)* & $1.396 \pm 0.010$ & $1.410 \pm 0.010$ & $1.736 \pm 0.012$  \\
\midrule
PF ($n = 1000$) & $1.531 \pm 0.022$ & $1.877 \pm 0.058$ & $14.690 \pm 0.312$  \\
\bottomrule
\end{tabular}

*Methods where the reported hyperparameters were found via grid search (see Appendix \ref{sec:grid}).
\end{table}

\begin{table}[ht]
\caption{RMSEs on the stochastic Lorenz attractor ($\alpha = 20$). Results show the average RMSE over 100 MC simulations with 95\% confidence intervals. RK4 indicates 4th order Runge-Kutta method. Euler indicates Euler's method. GRW indicates a Gaussian random walk.}
\label{experiment_2_q20}
\centering
\renewcommand{\arraystretch}{1.0} %
\begin{tabular}{lccc}
\toprule
\textbf{Method} & \textbf{RMSE (RK4)} & \textbf{RMSE (Euler)} & \textbf{RMSE (GRW)} \\
\midrule
EKF (original $\bm{Q}_{t-1}$) & $0.852 \pm 0.016$ & $1.193 \pm 0.027$ & $2.294 \pm 0.045$ \\
EKF (optimized $\bm{Q}_{t-1}$)* & $\bm{0.838 \pm 0.012}$ & $\bm{1.000 \pm 0.013}$ & $\bm{1.558 \pm 0.012}$ \\
IEKF ($K = 5$) & $0.852 \pm 0.016$ & $1.193 \pm 0.027$ & $2.294 \pm 0.045$ \\
\midrule
IMAP (Adadelta)* & $5.795 \pm 0.421$ & $7.244 \pm 0.468$ & $10.490 \pm 0.288$ \\
IMAP (Adam)* & $0.987 \pm 0.028$ & $1.183 \pm 0.015$ & $1.889 \pm 0.010$ \\
IMAP (Adagrad)* & $0.981 \pm 0.011$ & $1.148 \pm 0.014$ & $1.813 \pm 0.019$  \\
IMAP (RMSprop)* & $0.991 \pm 0.011$ & $1.127 \pm 0.017$ & $1.781 \pm 0.016$ \\
IMAP (Gradient Descent)* & $\bm{0.841 \pm 0.011}$ & $\bm{1.013 \pm 0.015}$ & $\bm{1.587 \pm 0.010}$  \\
\midrule
UKF (original $\bm{Q}_{t-1}$) & $2.000 \pm 0.034$ & $2.049 \pm 0.038$ & $4.293 \pm 0.092$  \\
UKF (optimized $\bm{Q}_{t-1}$)* & $1.428 \pm 0.011$ & $1.445 \pm 0.012$ & $1.734 \pm 0.014$  \\
\midrule
PF ($n = 1000$) & $1.612 \pm 0.033$ & $1.698 \pm 0.034$ & $13.535 \pm 0.474$  \\
\bottomrule
\end{tabular}

*Methods where the reported hyperparameters were found via grid search (see Appendix \ref{sec:grid}).
\end{table}

In Table \ref{supp_4_table}, we show the affect of modifying the number of gradient steps. In Table \ref{experiment_2_q1}, Table \ref{experiment_2_q5}, and Table \ref{experiment_2_q20} we show additional results for the stochastic Lorenz attractor for $\alpha = 1, 5, 20$, respectively. These additional results show both the original EKFs and UKFs and the EKFs and UKFs with $\bm{Q}_{t-1}$ optimized by grid search. For the optimized EKFs and UKFs, we test 500 $\bm{Q}_{t-1}$ matrices and report the best one. For the Implicit MAP Filters, we test 287 in total (35 Implicit MAP Filters are gradient descent). Again, we use 5 separate Monte Carlo simulations to select the Implicit MAP Filters to report.

The results are consistent with Table \ref{experiment_2_table} in Section \ref{sec:lorenz}. In all cases the EKF with optimized $\bm{Q}_{t-1}$ is statistically equivalent in performance to the Implicit MAP Filter with gradient descent. Notably, the EKF has access to a larger search space. Further, the stochastic Lorenz attractor is relatively convex, which suits the EKF, but the EKF diverges in highly nonlinear systems. The Implicit MAP Filter does not have this issue, working well in both Section \ref{sec:toy} and Section \ref{sec:lorenz}.

\subsubsection{Yearbook}\label{sec:extended3}

The yearbook dataset consists of 37,921 frontal-facing American high school yearbook photos from 1905 - 2013 from 128 high schools in 27 states. Each image is resized to a $32 \times 32 \times 1$ grayscale image and paired with a binary label $y$, representing the student’s gender. The years 1905 - 1930 only have 20 years with images available and years with as little as one photo available, making them not ideal for training and testing. Thus, we use these 20 years as a single pre-training dataset of 886 images. 

For all five methods we test, we use a 4-layer convolutional neural network (CNN). Each convolutional layer has a kernel size of $3 \times 3$, stride of $1 \times 1$, same padding, 32 output channels, ReLU activation,
and a 2D max pool layer with kernel size $2 \times 2$. We use stochastic gradient descent (SGD) with a fixed learning rate of $10^{-3}$, the pre-training dataset of 886 images, and a batch size of 64 to train the initial network for 200 steps. Since the results we report are the average over 10 random seeds, we pre-train the initial network 10 different times in exactly this fashion. This is supposed to resemble samples from some approximate initial starting distribution over neural network weights.

From 1931 - 2010, we take 32 randomly chosen images as a training set and 100 randomly chosen images as a test set at every time step. From 1931 - 1970, we take an additional 16 images as a validation set. In the main paper, we report test accuracy of the configurations with the best average validation performance from 1931 - 1970. The years 1941 and 2006 only had 93 and 70 images available, respectively. Thus, 1941 has a test set size of 45 and 2006 has a test set size of 38. All other training sets, validation sets, and test sets are exactly as described.

For the static weights approach, we do not do any additional training after the 200 steps of SGD over the pre-training dataset. We report the accuracy of those weights from 1931 - 2010 without any adaptation.

The direct fit approach uses 1000 full batch steps of the Adam optimizer at a fixed learning rate of $10^{-3}$ at every time step. This resembles what a practitioners might do to fit the training data. Instead of re-training from scratch, they would likely use the weight initialization from the previous time step. This inherently resembles the Implicit MAP filtering approach we propose here, but we are misspecifying the number of gradient steps by using 1000. 1000 gradient steps, or optimization to near convergence, assumes that the measurement noise is close to $\bm{0}$ at every time step. By using this overfitting Implicit MAP Filter, which we call direct fit, we are trying to show why it is important to appropriately define the \emph{implicit} filtering equations. It is naive to not consider the number of gradient steps as an important hyperparameter when working with time-varying objectives.

For the particle filter (PF) and variational Kalman filter (VKF), we explicitly specify a transition distribution $p(\rvw_t|\rvw_{t-1}) = \mathcal{N}(\rvw_t|\rvw_{t-1}, \sigma^2\bm{I})$ where $\sigma^2 \in \{0.1, 0.05, 0.01, 0.005, 0.001\}$. We run both of these methods exactly as described in Appendix \ref{sec:methods}.

Finally, we test five configurations of Adam for the Implicit MAP Filter where all hyperparameters were set to standard settings ($\eta = 0.001, \beta_1 = 0.9, \beta_2 = 0.999$) and the number of steps $K$ was selected from the set $\{1, 10, 25, 50, 100\}$. We use the validation set from the first 40 years to select the number of steps to report in the main paper. 

Figures 6, 7, and 8 show the classification accuracy of every IMAP filter, PF, and VKF, respectively. For Adam, performance was maximized by $K = 50$ and $K = 100$. The static weights case can be seen as an Implicit MAP Filter with $\bm{0}$ process noise over a transition function that is the identity function. This approach is a clear misspecification. 

The particle filters were generally only marginally better than the static weights case. This is due to a poorly defined proposal distribution, since it is hard to imagine that we could find the true transition distribution via grid search. 

The VKF saw better performance than the PF for nearly all configurations tested. This is due to the fact that we are explicitly optimizing the network weights. However, by explicitly regularizing the network instead of implicitly regularizing via early stopping, we do not match the performance of the Implicit MAP Filter or even the direct fit approach. This is likely due to the fact that the transition distribution is far from correctly specified. Again, we cannot imagine that we could find the correct transition distribution via grid search.

\newpage

\begin{figure}[h]
\label{yearbook_adam}
\begin{center}
\includegraphics[width=\textwidth]{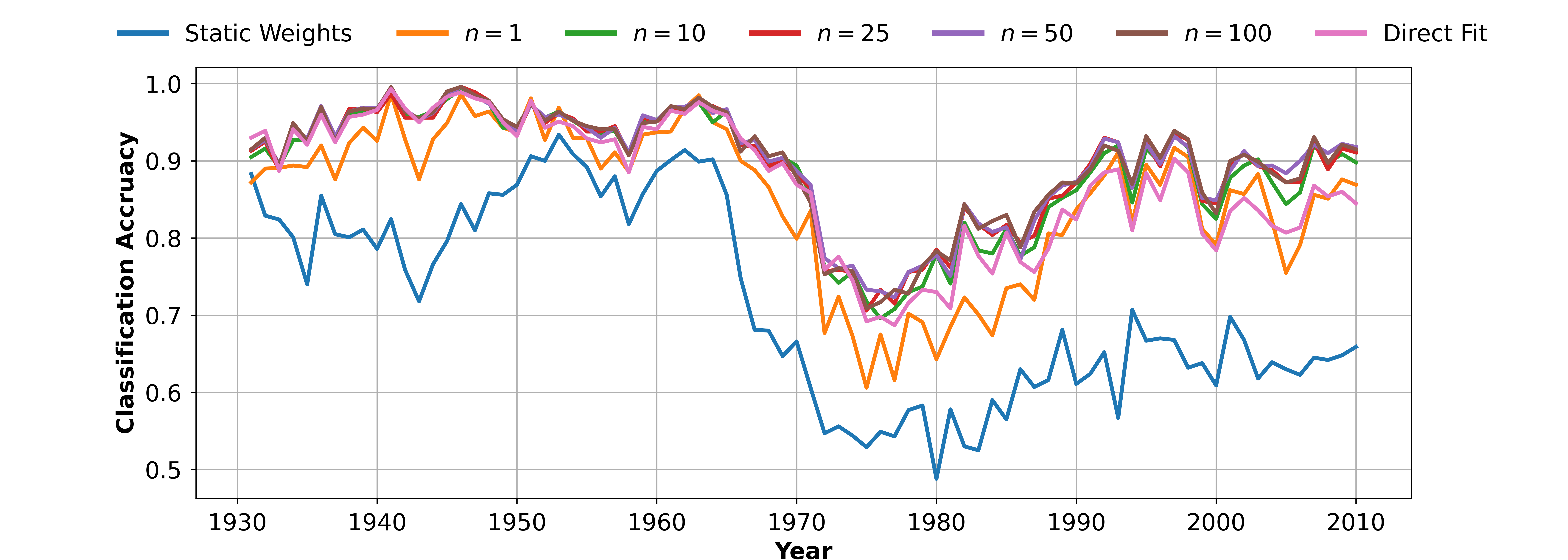}
\end{center}
\caption{Yearbook dataset filtering results using standard Adam hyperparameters over number of steps $K \in \{1, 10, 25, 50, 100\}$. We also plot the static weights case and direct fit case as comparative baselines.}
\end{figure}

\begin{figure}[h]
\label{yearbook_pf}
\begin{center}
\includegraphics[width=\textwidth]{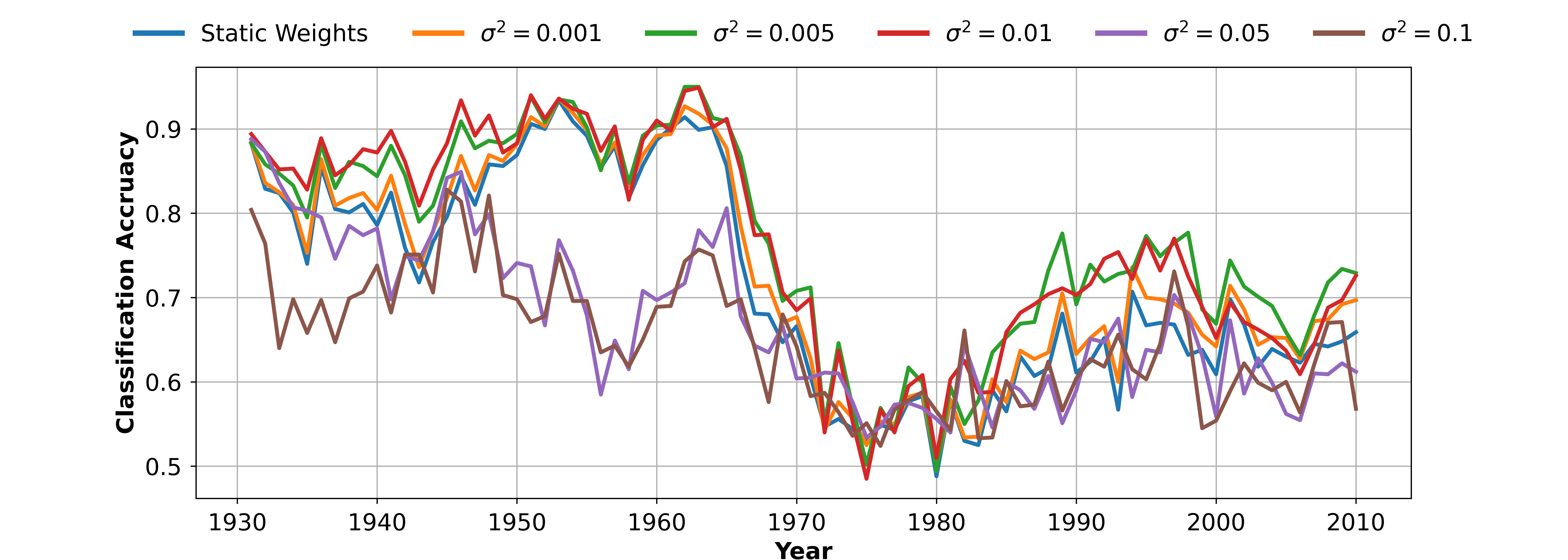}
\end{center}
\caption{Yearbook dataset filtering results using a particle filter with transition $\sigma^2 \in \{0.1, 0.05, 0.01, 0.005, 0.001\}$. We plot the static weights case as a comparative baselines.}
\end{figure}

\begin{figure}[h]
\label{yearbook_vkf}
\begin{center}
\includegraphics[width=\textwidth]{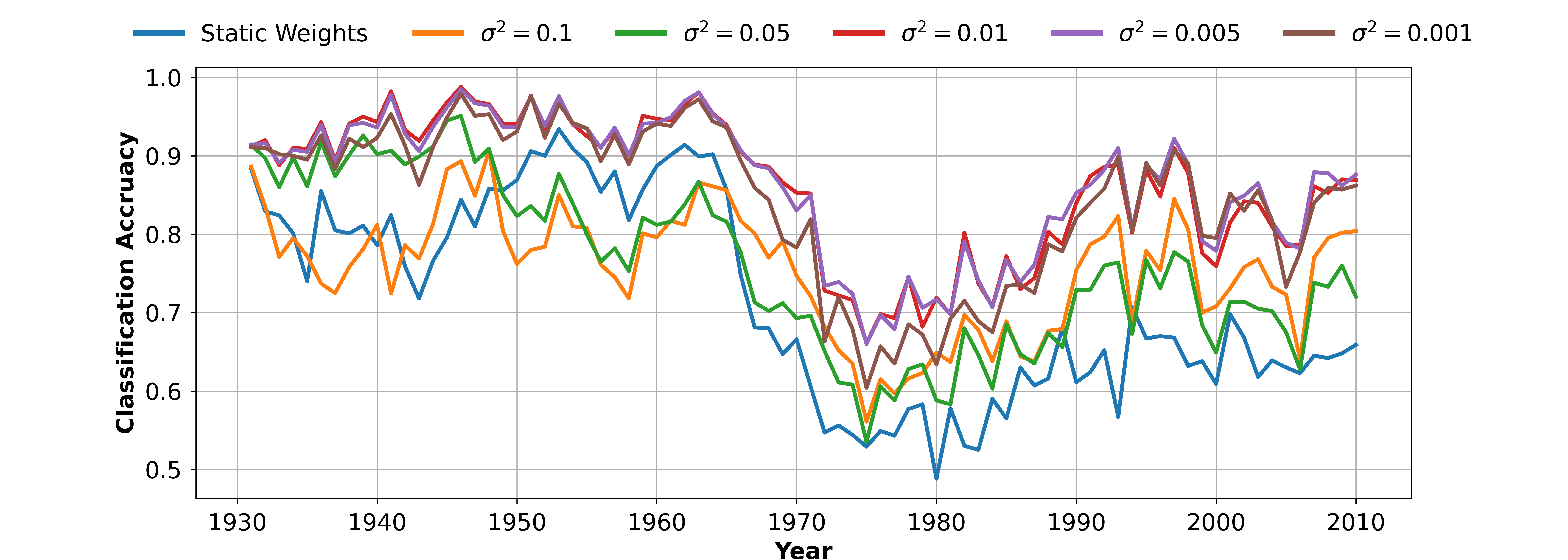}
\end{center}
\caption{Yearbook dataset filtering results using a variational Kalman filter with transition $\sigma^2 \in \{0.1, 0.05, 0.01, 0.005, 0.001\}$. We plot the static weights case as a comparative baselines.}
\end{figure}

\end{document}